\documentclass{article}
\ifdefined\pdfminorversion\pdfminorversion=7\fi 

\PassOptionsToPackage{numbers,compress}{natbib}
\usepackage[main,final]{neurips_2026}

\usepackage[utf8]{inputenc} 
\usepackage[T1]{fontenc}    
\usepackage{url}            
\usepackage{amsmath}        
\usepackage{booktabs}       
\usepackage{amsfonts}       
\usepackage{amssymb}        
\usepackage{nicefrac}       
\usepackage{microtype}      
\usepackage{xcolor}         
\usepackage{algorithm}      
\usepackage{algpseudocode}  

\usepackage{multirow}   
\usepackage{graphicx}   
\usepackage{makecell}
\usepackage{tabularx}
\usepackage{listings}
\usepackage{needspace}
\usepackage{placeins}
\usepackage{hyperref} 
\hypersetup{hidelinks,pdftitle={Automated Reformulation of Robust Optimization via Memory-Augmented Large Language Models},
  pdfauthor={Jinbiao Chen, Shuang Jin, Guoyun Zhang, Junyu Zhang, Guanyi Wang, Hanzhang Qin}}
\lstdefinestyle{promptbox}{
  basicstyle=\small\ttfamily,
  breaklines=true,
  breakatwhitespace=false,
  frame=single,
  framerule=0.5pt,
  backgroundcolor=\color{gray!8},
  xleftmargin=4pt,
  xrightmargin=4pt,
  columns=fullflexible,
  keepspaces=true,
}

\title{Automated Reformulation of Robust Optimization via Memory-Augmented Large Language Models}

\author{
Jinbiao Chen$^{1}$,
Shuang Jin$^{2}$,
Guoyun Zhang$^{3,4}$,\\
\bfseries Junyu Zhang$^{1}$,
Guanyi Wang$^{1,*}$,
Hanzhang Qin$^{1,3,}$\thanks{Corresponding Authors.}\\
$^1$Department of Industrial Systems Engineering and Management,\\
National University of Singapore\\
$^2$Department of Data and Systems Engineering, The University of Hong Kong\\
$^3$Institute of Operations Research and Analytics, National University of Singapore\\
$^4$Agency for Science, Technology and Research (A*STAR)\\
\texttt{bill.cjb@nus.edu.sg},~\texttt{jinsh7@hku.hk},~\texttt{zhang.guoyun@u.nus.edu}\\
\texttt{junyuz@nus.edu.sg},~
\texttt{guanyi.w@nus.edu.sg},~
\texttt{hzqin@nus.edu.sg}
}

\begin{document}

\maketitle

\begin{abstract}

  Robust optimization (RO) provides a principled framework for decision-making under uncertainty, but its practical use is often limited by the need to manually reformulate uncertain optimization models into tractable deterministic counterparts. Recent large language models (LLMs) have been shown promising for automating optimization formulation, yet RO reformulation remains challenging because it requires precise multi-step reasoning and mathematically consistent transformations. To facilitate systematic evaluation of LLM-based reformulation, for which no dedicated benchmark currently exists, we develop AutoRO-Bench, a benchmark featuring an automated data generation pipeline for the core RO reformulation task and a curated dataset for the RO application task. To address the reformulation challenge, we propose Automated Reformulation with Experience Memory (AutoREM), a tuning-free memory-augmented framework that autonomously builds a structured textual experience memory by reflecting on past failed trajectories through a tailored offline adaptation procedure. AutoREM requires neither domain-specific expert knowledge nor parameter updates, and the resulting memory readily transfers across different base LLMs. Experimental results show that AutoREM consistently improves the accuracy and efficiency of RO reformulation across in-distribution datasets, out-of-distribution datasets, and diverse base LLMs.
\end{abstract}

\section{Introduction}

\begin{figure}[t]
    \centering
    \includegraphics[width=1\linewidth]{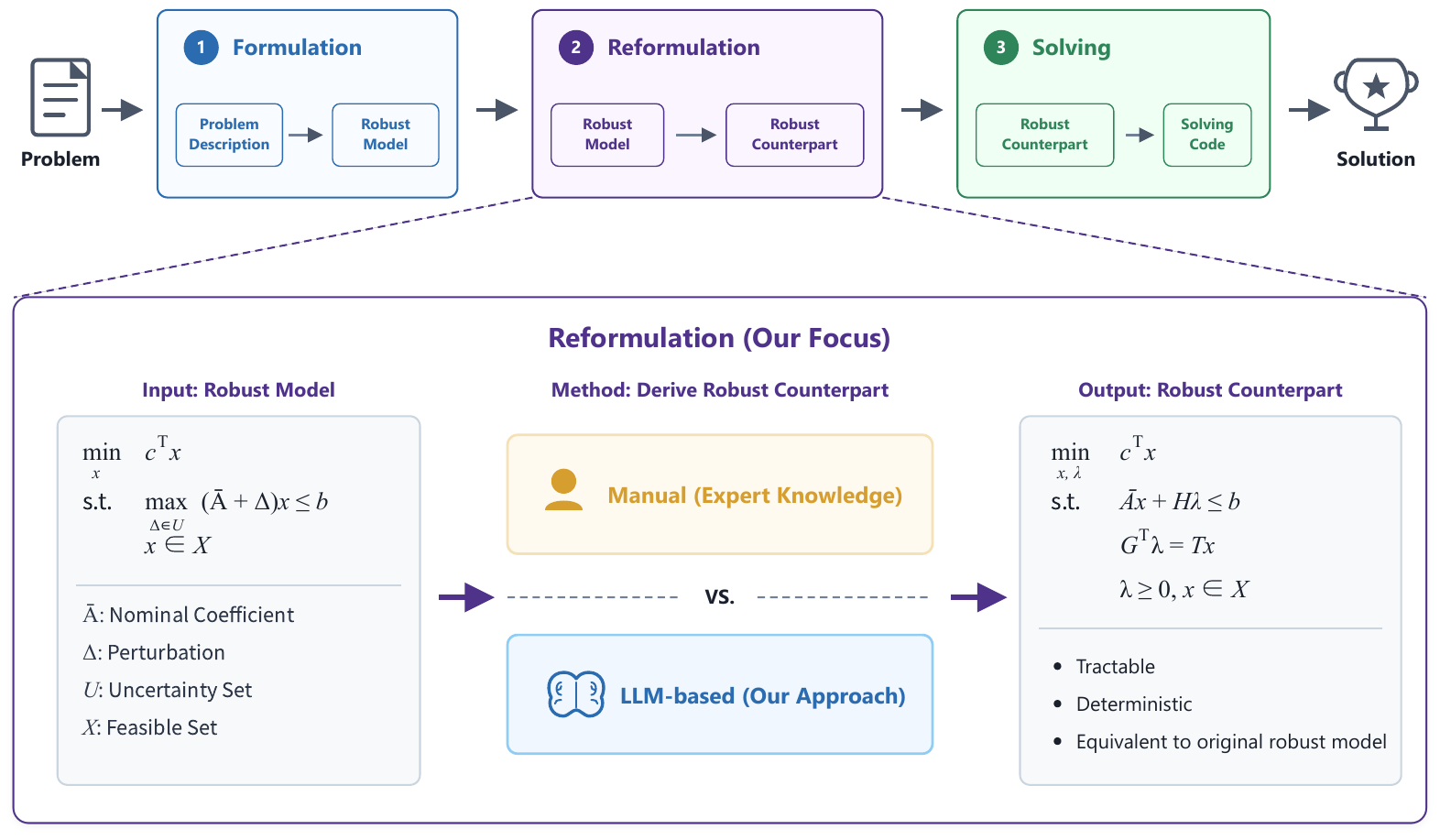} 
    \caption{The robust optimization pipeline and our focus on automated reformulation.}
    \label{fig:llm4ro}
\end{figure}

Robust optimization (RO) has been widely studied for producing reliable solutions under uncertainty \citep{bandi2012tractable, bertsimas2004price, ben-tal1999robust, ben-tal2000robust}. As illustrated in Figure \ref{fig:llm4ro}, applying RO typically follows a three-stage pipeline of formulation, reformulation, and solving. Recently, large language models (LLMs) have demonstrated strong reasoning and coding capabilities \citep{guo2025deepseekr1, hu2025openreasonerzero}, which greatly advances the automation of the solving stage and shows promise for automating the formulation stage \citep{xiao2025survey}. However, the reformulation stage remains a critical bottleneck: mathematically transforming uncertain models into tractable deterministic counterparts requires rigorous manual derivations grounded in convex analysis and duality theory, which severely limits the scalability of RO applications. This work, therefore, focuses on automating this reformulation stage with an LLM-based approach.

Recent methods for enhancing LLM reasoning are broadly divided into parameter-tuning and tuning-free approaches. Parameter-tuning methods, such as RL-based post-training, suffer from significant sample and computation inefficiency. They often require thousands of expensive rollouts and substantial computation to fit new tasks. Recent studies \citep{he2025rewarding, shao2025spurious, song2025outcomebased, yue2025does} further suggest that RL post-training primarily induces distribution sharpening rather than enabling genuinely new capabilities. By contrast, memory-augmented tuning-free methods \citep{hu2025memory, zhang2025survey} offer a highly scalable paradigm: they allow models to integrate interpretable and reusable knowledge from past trajectories at runtime, bypassing costly parameter updates and per-instance test-time search. This is uniquely suited for RO reformulation, where algebraic derivation and duality transformation patterns frequently recur across instances.

However, applying existing memory-augmented frameworks \citep{zhang2026agentic, ouyang2026reasoningbank} to RO reformulation typically suffers from two fundamental limitations: an additive-only paradigm and a lack of quality verification. As RO derivations must be mathematically exact and a single flawed memory trace can corrupt the full reasoning, memory quality can significantly outweigh quantity for RO reformulation capability. To systematically evaluate this capability, we introduce AutoRO-Bench, a dedicated benchmark with verified ground-truth solutions that comprises an automated data generation pipeline for the core reformulation task and a curated dataset for the application task. To overcome the limitations of existing memory-based methods, we propose Automated Reformulation with Experience Memory (AutoREM), a tuning-free framework with a tailored offline adaptation phase. AutoREM achieves fine-grained memory management through unit-level experiences (ULE) and structured memory operators (SMO), and ensures high-quality memory updates through dual-check commit (DCC) and validation-based acceptance (VBA).

In summary, this paper makes the following contributions. (1) We propose AutoRO-Bench, the first benchmark for systematically evaluating LLMs' RO reformulation capability. It features two complementary tasks: the RO reformulation task that takes mathematical models as input to test derivation capability of reformulation, and the RO application task that takes language-described problems as input to assess end-to-end practical problem solving, adopting benchmark settings similar to previous LLM-based formulation studies. (2) We propose AutoREM, a tuning-free, memory-augmented framework for automated RO reformulation, featuring a textual experience memory that accumulates reusable reformulation knowledge without expert instruction. (3) We design a tailored offline adaptation phase with four key components (ULE, SMO, DCC, and VBA) to automatically construct and refine high-quality reformulation memory. (4) Experiments demonstrate that AutoREM consistently improves accuracy and efficiency across both in-distribution and out-of-distribution datasets, and diverse base LLMs.

\section{Related work}

\textbf{LLMs for optimization formulation.} Recent studies have extensively explored LLMs for optimization formulation via parametric fine-tuning \citep{chen2025solverinformed, ding2026orr1, huang2025orlm, jiang2025llmopt, xiao2026deepor, xie2025murka, zhou2026steporlma} and tuning-free methods \citep{ahmaditeshnizi2024optimus, astorga2025autoformulation, jiang2025droc, kong2025alphaopt, li2025solverllm, liang2026largescale, liu2025mmagent, liu2025optitree, liu2026mathmo, pan2024guiding, ramamonjison2022augmenting, wang2025bppsearch, wang2025ormind, xiao2024chainofexperts, zhang2024solving, zhang2025decision}. However, automated RO reformulation remains largely unexplored, as existing methods focus on problem modeling, while lacking the capability to convert uncertain models to tractable counterparts. The closest work by \citet{bertsimas2024robust} relies heavily on manually crafted, domain-expert prompts and evaluates on only seven case studies. In contrast, we first introduce a significantly broader benchmark for systematic evaluation, and then propose AutoREM to autonomously collect reusable reformulation experiences without expert curation. Interestingly, the auto-generated memory from AutoREM consistently outperforms the expert-written prompt in the experiments, suggesting that the LLM reflector distills more targeted guidance from the LLM reformulator's specific behaviors than static expert knowledge.

\textbf{Tuning-free LLM reasoning methods.} Since the introduction of Chain-of-Thought (CoT) \citep{gu2022large, wei2022chain}, a useful distinction has emerged between intra-instance test-time scaling \citep{alon2023selfrefine, cassano2023reflexion,besta2024graph, cao2023tree, fu2026deep, kang2025scalable, karan2026reasoning, qi2025mutual, wang2023selfconsistency} and cross-instance context adaptation. Unlike test-time scaling, which processes isolated instances without retaining lessons, context adaptation efficiently accumulates reusable knowledge across instances. Within this paradigm, approaches divide into prompt optimization \citep{agrawal2026gepa, choi2025system, wang2024promptagent, wang2025evolving, zhang2024prefer} and memory augmentation \citep{zhang2026agentic, ouyang2026reasoningbank, bae2024autoguide, liu2025contextual, shu2026remem, zhao2023expel}. Notably, state-of-the-art (SOTA) memory methods \citep{zhang2026agentic} utilizing offline adaptation have recently demonstrated superiority over leading prompt optimizers \citep{agrawal2026gepa}. While prompt optimization suffers from a brevity bias that discards details, memory methods maintain a structured repository to accumulate domain-specific knowledge without information loss \citep{hu2025memory, zhang2025survey}. However, applying existing memory methods to RO reformulation is problematic due to their monolithic additive mechanisms and lack of rigorous verification. In fact, in our experiments, we even observe slight performance degrade over the base LLM after applying the SOTA memory methods.
To address these gaps, we design a tailored offline adaptation procedure to automatically construct and refine high-quality reformulation memory.

\section{Preliminary}

\subsection{Reformulation of robust optimization}
\label{sec:uncertainty-sets}

A \emph{robust model}~\citep{ben-tal1999robust,BenTal1998,BenTal2009} requires the feasibility to every uncertainty realization, yielding infinitely many constraints that render it directly intractable. \emph{Reformulation} derives a \emph{robust counterpart} (RC) of the original problem, which is an equivalent finite-dimensional deterministic problem that can be handled by standard solvers. A robust model is defined by a nominal problem and an uncertainty set.

\textbf{Nominal LP.}
We focus on linear programs (LPs) as the nominal problem class, which is classical and well-studied in robust optimization. A nominal LP takes the form:
\begin{equation}
\label{eq:nominal-lp}
\min_{\mathbf{x}\in\mathcal{X}}\;\bar{\mathbf{c}}^{\!\top}\mathbf{x}
\quad\text{s.t.}\quad
\bar{\mathbf{a}}_i^{\!\top}\mathbf{x}\leq b_i,\;i=1,\dots,m,
\end{equation}
where $\mathbf{x}\in\mathbb{R}^n$ is the decision vector, $\bar{\mathbf{c}}\in\mathbb{R}^n$ the cost vector, $\bar{\mathbf{a}}_i\in\mathbb{R}^n$ the $i$-th constraint row, and $\mathcal{X}$ any additional deterministic constraints.

\textbf{Uncertainty set.}
An uncertainty set specifies admissible perturbations to the nominal coefficients. In general, they take the form $\tilde{\mathbf{c}}=\bar{\mathbf{c}}+\mathbf{P}^c\boldsymbol{\zeta}^c$ and $\tilde{\mathbf{a}}_i=\bar{\mathbf{a}}_i+\mathbf{P}_i\boldsymbol{\zeta}_i$, with $\boldsymbol{\zeta}^c\in\mathcal{U}_c$, $\boldsymbol{\zeta}_i\in\mathcal{U}_i$ and perturbation matrices $\mathbf{P}^c$, $\mathbf{P}_i$. We take $\mathbf{P}^c=\mathbf{P}_i=\mathbf{I}$, yielding the element-wise form $\tilde{c}_j=\bar{c}_j+\zeta_j^c$ and $\tilde{a}_{ij}=\bar{a}_{ij}+\zeta_{ij}$. This avoids the extraneous numerical entries in a general $\mathbf{P}_i$ and keeps the focus on the reformulation derivation itself. We consider three classical choices: box, budget, and polyhedral uncertainty set. Each \emph{row} (i.e., the objective function or a constraint) of the model may independently be assigned any of these uncertainty types. Together with the nominal LP~\eqref{eq:nominal-lp}, they define the robust model:
\begin{equation}
\label{eq:robust-model}
\min_{\mathbf{x}\in\mathcal{X}}\;\max_{\boldsymbol{\zeta}^c\in\mathcal{U}_c}
  (\bar{\mathbf{c}}+\boldsymbol{\zeta}^c)^{\!\top}\mathbf{x}
\quad\text{s.t.}\quad
\max_{\boldsymbol{\zeta}_i\in\mathcal{U}_i}(\bar{\mathbf{a}}_i+\boldsymbol{\zeta}_i)^{\!\top}\mathbf{x}\leq b_i,
\;i=1,\dots,m.
\end{equation}

\textbf{Robust counterpart (RC).}
Reformulation eliminates each inner maximization in~\eqref{eq:robust-model} by LP duality, yielding a finite set of linear constraints. For all three uncertainty sets above, the RC is an LP. Table~\ref{tab:uncertainty} summarizes their definitions and RC forms; full derivations can be found in Appendix~\ref{app:rc-derivations}.

\begin{table}[t]
\centering
\caption{Three uncertainty sets and their LP robust counterparts.}
\label{tab:uncertainty}
\resizebox{\textwidth}{!}{
    \renewcommand{\arraystretch}{1.5}
    \begin{tabular}{lll}
    \toprule
    \textbf{Uncertainty} & \textbf{Definition of $\mathcal{U}_i$} & \textbf{Robust Counterpart} \\
    \midrule
    Box & $|\zeta_j|\leq \delta_j, \forall j$ & $\bar{\mathbf{a}}_i^\top\mathbf{x}+\sum_j t_j\leq b_i; t_j\geq\pm\delta_j x_j, t_j\geq 0$ \\
    Budget & $|\zeta_j|\leq \delta_j, \forall j; \sum_j|\zeta_j|/\delta_j\leq\Gamma_i$ & $\bar{\mathbf{a}}_i^\top\mathbf{x}+\Gamma_i z_{i0}+\sum_j z_{ij}\leq b_i; z_{i0}+z_{ij}\geq\pm\delta_j x_j, z_{i0},z_{ij}\geq 0$ \\
    Polyhedral & $\mathbf{F}\boldsymbol{\zeta}_i\leq\mathbf{g}$ & $\bar{\mathbf{a}}^\top\mathbf{x}+\mathbf{g}^\top\boldsymbol{\lambda}\leq b; \mathbf{F}^\top\boldsymbol{\lambda}=\mathbf{x}, \boldsymbol{\lambda}\geq\mathbf{0}$ \\
    \bottomrule
    \end{tabular}
}
\end{table}

\subsection{Memory-augmented LLM methods}
\label{sec:preliminary}

Let $\theta$ denote the parameters of an LLM. In standard generation, the model receives an input problem $q$ and a task prompt $p$, generating a response $y \sim P_\theta(y \mid p, q)$. To enhance reasoning capabilities without updating $\theta$, memory-augmented methods \citep{hu2025memory, zhang2025survey} introduce a structured repository of reusable external memory $\mathcal{M}$ 
so that the model can draw on accumulated past experience at inference time.

\textbf{Memory retrieval.}
Rather than a simple string, $\mathcal{M}$ is typically a structured textual object.
It follows two retrieval paradigms. In \emph{explicit retrieval}, an external module pre-selects a relevant subset $\mathcal{M}_\text{sub}$ (e.g., via semantic similarity \citep{ouyang2026reasoningbank}), yielding $y \sim P_\theta(y \mid p, \mathcal{M}_\text{sub}, q)$.
Conversely, \emph{implicit retrieval} supplies the entire $\mathcal{M}$ directly to the context window \citep{zhang2026agentic}, relying on the LLM's self-attention to dynamically filter information, yielding $y \sim P_\theta(y \mid p, \mathcal{M}, q)$.

\textbf{Memory formation.}
Memory is constructed by distilling generalizable insights from past trajectories into structured, editable entries.
This proceeds in two modes. \emph{Online adaptation} continuously updates $\mathcal{M}$ during inference, enabling continual learning on new tasks.
Alternatively, \emph{offline adaptation} constructs $\mathcal{M}$ on a pre-deployment dataset; given abundant data, this enables thorough reflection to build a high-quality memory without compromising inference efficiency.

\section{Methodology}

\subsection{AutoRO-Bench: automated RO reformulation benchmark}
\label{sec:benchmark}

To comprehensively evaluate LLM-based methods for robust optimization (RO) reformulation, we introduce AutoRO-Bench, comprising two complementary tasks that cover different input modalities: the \emph{RO reformulation task}, which takes formal mathematical models expressed in \LaTeX\ as input, and the \emph{RO application task}, which takes natural-language problem descriptions as input.

\textbf{RO reformulation task.}
The RO reformulation task tests the core mathematical challenge: given a formal RO model in \LaTeX, derive its deterministic robust counterpart. Each instance consists of two components: a \LaTeX\ problem description as the LLM input $q$, and a ground-truth optimal value $f^*$  as the evaluation target. The problem is a robust LP model~\eqref{eq:nominal-lp} with uncertain coefficients as in~\eqref{eq:robust-model}. Although the symbolic robust counterpart cannot be produced automatically, $f^*$ is obtained by directly solving the robust model via the \texttt{RSOME} solver~\citep{chen2023rsome,chen2020robust}.

To generate such verified instances at scale, we propose an automated generation pipeline (Algorithm~\ref{alg:data_gen} in Appendix~\ref{appx:data_gen}). It takes as input the number of decision variables $n$, a target number of valid instances $N$, a candidate uncertainty type set $\mathcal{U}$ (instantiated with box, budget, and polyhedral), and a template set $\mathcal{T}$ (instantiated with $8$ diverse templates $T_{b_0 b_1 b_2}$ parameterized by three binary dimensions to encourage notation-invariant learning, see Appendix~\ref{appx:template}). For each candidate instance, the pipeline proceeds through five steps: \textbf{(1)} generate a nominal LP by sampling objective coefficients $\mathbf{c}$ and optimization direction, and constructing a sparse feasible constraint polytope $(\mathbf{A}, \mathbf{b})$ with a guaranteed interior point $\mathbf{x}_0$ via the \textsc{FeasiblePolytope} algorithm (Appendix~\ref{appx:feasiblelp}). \textbf{(2)} Assign an uncertainty type from $\mathcal{U}$ to each row. \textbf{(3)} For each uncertain row, randomly select a sparse subset $\mathcal{S}_i$ of coefficients as the uncertain support, then generate the corresponding uncertainty parameters. The box and budget types use dedicated sampling procedures (Appendix~\ref{appx:datastep3}), while polyhedral type employs \textsc{FeasiblePolytope} to construct a sparse uncertainty polytope. \textbf{(4)} Solve via \texttt{RSOME} and discard infeasible, unbounded, or degenerate instances. \textbf{(5)} Render the accepted instance into \LaTeX\ using the selected template(s). Each output record contains the \LaTeX\ problem description, optimal solution $\mathbf{x}^*$, and optimal value $f^*$.

\textbf{RO application task.}
The RO application task evaluates end-to-end performance on realistic RO problem solving. Its natural-language problem description setting is aligned with existing LLM-based optimization formulation benchmarks, enabling direct and comprehensive comparison with those methods. To support this task, we curate the \emph{RO application dataset} of $32$ instances. We first select $16$ nominal problems from two established benchmarks  with $8$ from IndustryOR~\citep{huang2025orlm} and $8$ from OptiBench~\citep{yang2025optibench}, covering both LP and mixed-integer linear program (MILP). For each nominal problem, we append a natural-language description of the uncertain parameters and the robustness requirement to the original problem statement, constructing a complete RO specification in plain text (see Appendix~\ref{appx:app_format}).

We generate two RO instances per nominal problem: the first uses the same parameter sampling procedure as the RO reformulation task and  the second uses practically-motivated parameters following the RO literature~\citep{bandi2012tractable, ben-tal2000robust}. Unlike the first procedure where uncertainty parameters are given as explicit numerical values, the second leaves some parameters as unevaluated expressions involving operations such as ratios or square roots over problem quantities. For example, deviations may be given as proportions of nominal coefficients, or polyhedral bounds may involve the square root of the uncertain coefficient count. These expressions must be preserved in the reformulation output rather than evaluated to a single number. Full parameter details are given in Appendix~\ref{appx:poly_types}. Since each instance is highly tailored to its nominal problem, all $32$ instances were independently verified by three operations research domain experts. Dataset statistics are summarized in Appendix~\ref{appx:app_stats}.

\textbf{Evaluation protocol.}
The benchmark evaluates reformulation quality by solver verification: a model receives problem description $q$ and outputs the robust counterpart; the output is translated into executable code and run by a standard LP/MILP solver to obtain $\hat{f}$, which is compared against the ground-truth $f^*$ stored in each data point. An instance is correctly reformulated if $\hat{f}$ matches $f^*$ within a numerical tolerance, providing an automatic, solver-grounded evaluation signal agnostic to the syntactic form of the output. Concrete examples of both tasks are provided in Appendix~\ref{appx:examples}.

This paper employs a two-stage pipeline with a \emph{reformulator} and a \emph{coder}, each an LLM with distinct roles. The reformulator, parameterized by $\theta$, takes problem description $q$ and task prompt $p$ as input and generates the robust counterpart $y \sim P_\theta(y \mid p, q)$ or $y \sim P_\theta(y \mid p, \mathcal{M}, q)$ with experience memory $\mathcal{M}$. The coder, parameterized by $\phi$, takes $y$ and a coding prompt $p'$ as input and generates executable solver code $c \sim P_\phi(c \mid p', y)$; running $c$ yields $\hat{f}$. We use cost-efficient flagship LLMs as the reformulator and fix a substantially stronger but also more expensive model as the coder. Coding from a correct reformulation is mechanical, so fixing a strong coder isolates reformulation quality as the sole variable.

\subsection{AutoREM: automated reformulation with experience memory}

\begin{figure}[t]
    \centering
    \includegraphics[width=1\linewidth, height = 7.5cm]{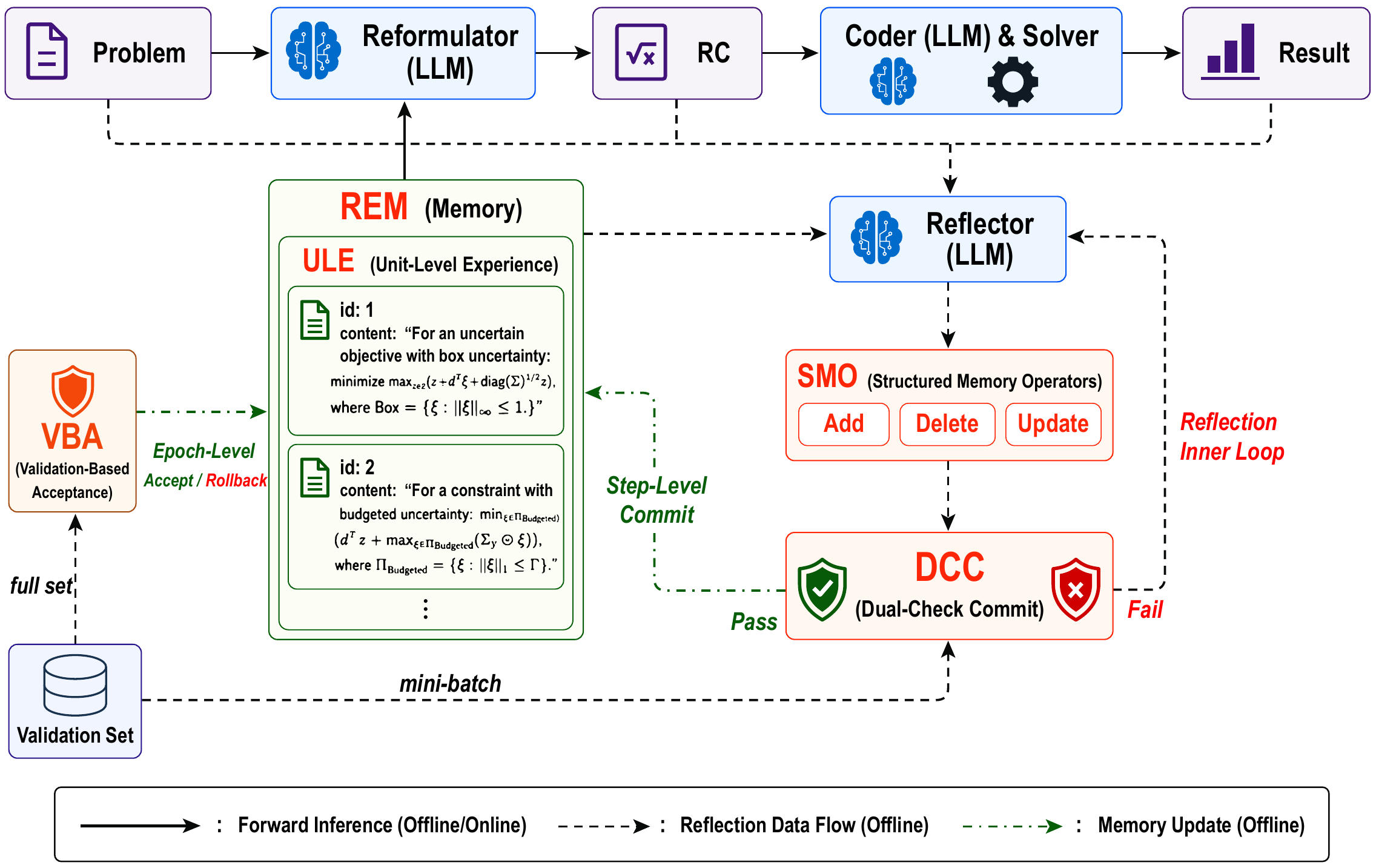}
    \caption{Overview of the AutoREM pipeline.}
    \label{fig:method}
\end{figure}

We propose AutoREM, a tuning-free LLM framework augmented with a reformulation experience memory (REM). As shown in Figure~\ref{fig:method}, it operates through two interconnected phases: \emph{offline adaptation} that constructs a high-quality memory $\mathcal{M}$ on a dedicated offline dataset before deployment, and \emph{online inference} that leverages the resulting $\mathcal{M}$ to guide the reformulator at test time.

\textbf{Online inference.} During inference, we employ the implicit retrieval paradigm: the entire memory $\mathcal{M}$ is provided directly in the reformulator's context, formulated as $y \sim P_\theta(y \mid p, \mathcal{M}, q)$, allowing the model to dynamically attend to relevant past derivation patterns without any external retrieval.

\textbf{Offline adaptation.} The quality of online inference depends on the mathematical reliability of $\mathcal{M}$. In RO reformulation where derivations must be exact, even a single erroneous memory entry can corrupt reasoning, making quality far more critical than quantity. This motivates a dedicated offline adaptation phase rather than relying on manual curation or expensive test-time search. Crucially, the AutoRO-Bench pipeline provides rich training instances at low cost, making data-driven memory construction feasible and scalable.

\subsection{Offline adaptation of reformulation experience memory}
\label{sec:offline}

The offline adaptation phase constructs and refines REM through four tightly coupled components organized around two design principles. The first principle is fine-grained memory management: unit-level experience (ULE) addresses the representation dimension by organizing knowledge at the granularity of individual uncertain constraint rows, making each entry independently reusable; structured memory operators (SMO) address the editing dimension by providing atomic add, update, and delete operations for precise, interpretable memory modification. The second principle is high-quality memory verification: dual-check commit (DCC) operates at the step level, stress-testing each proposed update against a targeted validation batch before committing; validation-based acceptance (VBA) operates at the epoch level, evaluating the full validation set and rolling back any harmful accumulated changes.

\textbf{Structured representations.}
Following prior memory-augmented methods \citep{zhang2026agentic, ouyang2026reasoningbank}, $\mathcal{M}$ is organized as an ordered list of  independently editable entries. Each entry has two fields: a unique integer \texttt{experience\_id} for identification, and a free-form \texttt{content} field encoding reformulation knowledge for one minimal independent reformulation unit. The Reformulator and Reflector outputs adopt a unified JSON format. The Reformulator returns \texttt{reasoning}, \texttt{matched\_experience\_ids} (the entries used, supporting fine-grained attribution), and \texttt{final\_answer} (the robust counterpart in \LaTeX). The Reflector returns \texttt{reasoning} and \texttt{final\_answer} (a list of memory operators). An external controller parses these outputs and applies the operations. See full prompts in Appendix~\ref{appx:prompts}.

\textbf{Overall flow.}
Adaptation proceeds for $E$ epochs of $S$ steps each. At each step, a training instance $q_s$ is evaluated under $\mathcal{M}$. If it is already correct, then the step is skipped. Otherwise, the system enters an \emph{inner reflection loop} of up to $K$ iterations, where a Reflector LLM proposes SMO edits on a temporary copy $\hat{\mathcal{M}}$ that targets ULE entries while leaving the committed $\mathcal{M}$ untouched. Once $q_s$ becomes correct under $\hat{\mathcal{M}}$, the DCC gate validates it on a locked batch of validation samples: a full pass commits $\hat{\mathcal{M}}$ to $\mathcal{M}$, while any regression feeds the failure details back into the next Reflector call. If no commit occurs within $K$ iterations, $\hat{\mathcal{M}}$ is discarded. After each step, the controller records a usage-success or usage-failure event on every matched experience, allowing the Reflector to perceive per-experience quality accumulated over training. At epoch end, VBA evaluates the full validation set. If the current accuracy is below the best seen so far across all epochs (tracked as $\mathcal{M}^*$), $\mathcal{M}$ is rolled back to $\mathcal{M}^*$ before next epoch. See complete procedure in Algorithm~\ref{alg:offline}, Appendix~\ref{appx:offline_alg}.

\textbf{Unit-level experience (ULE).}
ULE defines the granularity of memory entries: each entry targets one minimal independent reformulation unit, making it self-contained and independently applicable to any structurally similar unit in a new problem. This fine decomposition enables more precise attribution and flexible reuse of reformulation knowledge than coarser, problem-level experiences. In practice, ULE is realized by simply adding the phrase ``minimal independent reformulation unit'' to the Reflector's prompt. The Reflector itself decides what counts as such a unit for each problem and distills experience at that level.

\textbf{Structured memory operators (SMO).}
The Reflector outputs a JSON list of memory operations, each of three types: \textsc{add} (insert a new entry), \textsc{update} (revise an existing entry's content), or \textsc{delete} (remove an entry), each specifying the target \texttt{experience\_id}. An external controller parses this JSON output and applies the corresponding operations to the working copy $\hat{\mathcal{M}}$, while leaving the committed $\mathcal{M}$ untouched. The Reflector also receives the full trace of prior operations and outcomes of the inner loop, enabling it to diagnose why earlier edits failed and propose targeted corrections.

\textbf{Dual-check commit (DCC).}
Committing after a single training fix risks silent regression on previously correct instances. When $q_s$ first becomes correct under $\hat{\mathcal{M}}$, a mini-batch of up to $B$ validation instances is selected from the validation set and locked for the remainder of the step. Selection prioritizes \emph{linked} instances in the validation set, whose prior correct answers are attributed to the modified experience entries, over the \emph{unlinked} ones, filling the remaining budget with currently correct unlinked instances. Linking is maintained via an experience-instance correspondence map, which the external controller updates by reading the \texttt{matched\_experience\_ids} field in the Reformulator's JSON output.  ULE's fine-grained unit-level decomposition makes this attribution more accurate. The locked batch is then evaluated under $\hat{\mathcal{M}}$. On DC pass, $\hat{\mathcal{M}}$ is committed to $\mathcal{M}$ and the correspondence map updated. On DC fail, the regression details (failing problem, reformulator output, and solver feedback) are returned to the Reflector as additional context, making DCC an active source of learning signal rather than a passive filter.

\textbf{Validation-based acceptance (VBA).}
DCC covers only a mini-batch of $B$ instances per step, which provides a fast but biased estimate of memory quality and can let locally beneficial yet globally harmful edits slip through. To obtain a more reliable assessment, VBA further evaluates the full fixed held-out validation set of size $|\mathcal{V}|$ at each epoch end and retains the best-ever snapshot as $\mathcal{M}^*$. If the epoch-end accuracy falls below the best recorded value, the current $\mathcal{M}$ is rolled back before the next epoch begins, preventing the accumulation of such harmful edits over time.

\section{Experiments}
\label{sec:experiments}

\subsection{Experimental settings}
\label{sec:experimental-settings}

\textbf{Tasks and datasets.}
We evaluate AutoREM on the two tasks of AutoRO-Bench, each with a distinct evaluation goal and a separate set of baselines.

For the RO reformulation task, we compare AutoREM against LLM-based reformulation methods using three datasets generated by our pipeline. The \emph{Random} dataset (64 instances) serves as the in-distribution benchmark, following the same generation procedure as the training data with 2--5 decision variables. The \emph{Hard} dataset (32 instances) is an out-of-distribution stress test where each instance is designed around observed failure modes of the base LLM (see Appendix~\ref{appx:hard_dataset}). The \emph{Large} dataset (32 instances) is a second out-of-distribution split that uses the same random procedure as \emph{Random} but scales the variable count to 6 and 7, probing generalization to larger problem dimensions.
Train, validation, and test instances are independently generated without reusing underlying problems or uncertainty-parameter sets. Split-independence and input-length details are provided in Appendix~\ref{appx:evaluation_protocol}.


For the RO application task, we instead compare AutoREM against LLM-based formulation methods, evaluating end-to-end RO problem solving from natural-language input on the 32-instance RO \emph{Application} dataset. Following the full RO solving pipeline (Figure~\ref{fig:llm4ro}), we first directly employ the same base LLM as a formulator to translate each natural-language problem description into a \LaTeX\ robust model, which is then passed to AutoREM as the reformulation component.


\textbf{Baselines.}
We compare AutoREM against two groups of baselines, all using the same base LLM. The datasets and code are publicly available\footnote{\url{https://github.com/Bill-CJB/AutoREM}} to support reproducibility.

\emph{LLM reformulation methods} operate on the RO reformulation task, taking a \LaTeX\ robust model as input; all use zero-shot CoT \citep{gu2022large} prompting and share the same fixed coder LLM:
(1)~\textbf{Base LLM};
(2)~\textbf{Max Thinking}: same base LLM with maximum reasoning effort, scaling test-time computation per instance;
(3)~\textbf{Expert Prompt} \citep{bertsimas2024robust}: the RO domain expert prompt;
(4)~\textbf{ReasoningBank} \citep{ouyang2026reasoningbank}: a SOTA online adaptation memory method, adapted to our offline setting;
(5)~\textbf{ACE} \citep{zhang2026agentic}: a SOTA offline adaptation memory framework, the most related competitor.
AutoREM, ACE, and ReasoningBank each receive $10\times8=80$ training exposures. Offline adaptation protocols are detailed in Appendix~\ref{appx:baseline_and_coverage}.

\emph{LLM formulation methods} address the RO application task end-to-end from natural-language input:
(6)~\textbf{AlphaOPT} \citep{kong2025alphaopt}: an experience-library-based formulation method;
(7)~\textbf{LEAN-LLM-OPT} \citep{liang2026largescale}: a retrieval-augmented-generation-based agentic formulation framework;
(8)~\textbf{OptiTree} \citep{liu2025optitree}: a tree-search-based test-time scaling formulation method.

\textbf{Hyperparameter settings.}
All methods use DeepSeek-V4-Flash as the base LLM with default reasoning effort (high), maximum output tokens of 32K, and temperature 0 (greedy decoding). DeepSeek-V4-Pro is used as the fixed coder. For the offline adaptation phase, we run $E = 10$ epochs with $S = 8$ steps per epoch. The dual-check batch size is set to $B = 4$, the maximum inner reflection iterations to $K = 3$, and the validation set size to $|\mathcal{V}| = 64$.

\textbf{Metrics.}
We report two metrics to jointly assess reformulation quality and efficiency.
\emph{Accuracy} (percentage of instances whose reformulation yields an optimal objective value matching the ground truth within numerical tolerance) measures reformulation quality.
The objective-agreement criterion is $|\hat f-f^*|<10^{-4}$. Malformed outputs and solver failures count as incorrect.
\emph{Output Tokens} (average number of tokens generated by the reformulator per instance) measures reformulation efficiency: output tokens are the primary determinant of online inference latency, and output tokens are priced significantly higher than prompt tokens in API-based deployments, making this a practically important metric.
Although temperature is set to 0 (greedy decoding), LLM responses may still vary across API calls due to engineering-level non-determinism. We therefore run each method three times with the same frozen learned memory where applicable and report the mean $\pm$ sample standard deviation.
In all tables, \textbf{bold} indicates the best result and \underline{underline} indicates the second best.

\subsection{Experimental results}
\label{sec:reformulation-results}

\textbf{Comparison.}
AutoREM achieves the highest accuracy on the in-distribution \emph{Random} dataset (97.4\%), surpassing Expert Prompt by 4.7\%, Max Thinking by 6.8\%, and both SOTA memory methods (ReasoningBank, ACE) by more than 10\%.
It also attains the best accuracy and efficiency trade-off, generating 54\% fewer tokens than Max Thinking while delivering far higher accuracy.
Notably, both SOTA memory methods (ReasoningBank: 85.4\%, ACE: 87.0\%) fail to improve over the Base LLM (87.5\%), showing that on mathematically rigorous tasks, blindly accumulating experiences is counterproductive.
In contrast, AutoREM's selective adaptation converges to just 8 verified, non-redundant templates, demonstrating that high-quality experiences matter far more than quantity. See Appendices~\ref{appx:memory_comparison} and~\ref{appx:offline_curve} for the full qualitative comparison and adaptation dynamics.

\begin{table}[t]
  \centering
  \caption{Comparison of methods across the three datasets of the RO reformulation task.}
  \label{tab:main}
\resizebox{\textwidth}{!}{
    \begin{tabular}{lcccccc}
    \toprule
    \multirow{2}[2]{*}{Methods} & \multicolumn{2}{c}{Random (In-Distribution)} & \multicolumn{2}{c}{Hard (Out-of-Distribution)} & \multicolumn{2}{c}{Large (Out-of-Distribution)} \\
          & Accuracy/\%↑ & Output Tokens↓ & Accuracy/\%↑ & Output Tokens↓ & Accuracy/\%↑ & Output Tokens↓ \\
    \midrule
    Base LLM & 87.5 ± 5.4 & 7777 ± 162 & 70.8 ± 10 & 9026 ± 295 & 76.0 ± 3.6 & 11757 ± 135 \\
    Max Thinking & 90.6 ± 1.6 & 13857 ± 331 & \underline{83.3 ± 1.8} & 14902 ± 651 & 77.1 ± 3.6 & 22695 ± 1164 \\
    Expert Prompt & \underline{92.7 ± 3.3} & 8750 ± 200 & \underline{83.3 ± 3.6} & 7549 ± 101 & \underline{79.2 ± 9.5} & 13243 ± 1033 \\
    ReasoningBank & 85.4 ± 0.9 & 6575 ± 131 & 80.2 ± 4.8 & 8089 ± 701 & 71.9 ± 13.6 & 11774 ± 471 \\
    ACE   & 87.0 ± 6.3 & \textbf{4419 ± 56} & 81.3 ± 9.4 & \textbf{5238 ± 570} & 63.5 ± 4.8 & \textbf{6706 ± 531} \\
    AutoREM & \textbf{97.4 ± 0.9} & \underline{6386 ± 56} & \textbf{94.8 ± 3.6} & \underline{6944 ± 27} & \textbf{85.4 ± 7.1} & \underline{10929 ± 238} \\
    \bottomrule
    \end{tabular}%
  }
\end{table}

\textbf{Out-of-distribution generalization.}
AutoREM generalizes strongly to both out-of-distribution settings.
On the \emph{Hard} dataset, the advantage is most pronounced: all baselines drop substantially from their \emph{Random} scores (e.g., Expert Prompt: $-$9.4\%, ACE: $-$5.7\%), whereas AutoREM declines only marginally ($-$2.6\%), outperforming the best baseline by 11.5\% and the Base LLM by 24.0\%.
This robustness demonstrates that the learned memory genuinely improves the model's core reformulation capability rather than memorizing training-set patterns.
AutoREM also reduces output tokens by 23\% compared to the Base LLM.
On the \emph{Large} dataset, all methods exhibit a consistent accuracy drop, indicating that the bottleneck is long-context processing rather than reformulation capability.
AutoREM still leads at 85.4\% with consistent efficiency gains.

We further evaluate generalization across uncertainty structures and problem sources using the same frozen memory without re-adaptation. The three 32-instance datasets cover non-diagonal perturbation mappings (\emph{General $P_i$}), shared primitive uncertainty across rows (\emph{Coupled Row}), and textbook and published robust models independent of our generator (\emph{External}). AutoREM improves accuracy from 89.6\% to 90.6\%, from 76.0\% to 89.6\%, and from 92.7\% to 97.9\%, respectively. These results provide additional evidence that the trained memory transfers to unseen instances sharing relevant reformulation patterns, including textbook and published RO models beyond our generator (Appendix~\ref{appx:transfer-tests}).

\textbf{Cross-model universality.}
To evaluate transferability, we directly apply the memory trained on DeepSeek-V4-Flash to GPT-5.4 and Qwen3.6-Plus without re-adaptation.
These three models are high cost-efficiency flagship offerings from  DeepSeek, OpenAI, and Alibaba, providing broad coverage of model architectures and training paradigms.
As shown in Table~\ref{tab:basellm}, AutoREM consistently improves both accuracy and token efficiency across all three base LLMs, with accuracy gains of 6.3\% and 7.3\% on GPT-5.4 and Qwen3.6-Plus respectively.
This indicates that the reformulation experience memory encodes transferable knowledge that generalizes across heterogeneous base LLMs.

\begin{table}[t]
  \centering
  \caption{Comparison of methods across three base LLMs.}
  \label{tab:basellm}
\resizebox{\textwidth}{!}{
    \begin{tabular}{lcccccc}
    \toprule
    \multirow{2}[2]{*}{Methods} & \multicolumn{2}{c}{DeepSeek-V4-Flash} & \multicolumn{2}{c}{GPT-5.4} & \multicolumn{2}{c}{Qwen3.6-Plus} \\
          & Accuracy/\%↑ & Output Tokens↓ & Accuracy/\%↑ & Output Tokens↓ & Accuracy/\%↑ & Output Tokens↓ \\
    \midrule
    Base LLM & 87.5 ± 5.4 & 7777 ± 162 & 90.6 ± 3.1 & 2871 ± 78 & 81.2 ± 2.7 & 3448 ± 74 \\
    AutoREM & \textbf{97.4 ± 0.9} & \textbf{6386 ± 56} & \textbf{96.9 ± 0} & \textbf{2550 ± 21} & \textbf{88.5 ± 1.8} & \textbf{2697 ± 73} \\
    \bottomrule
    \end{tabular}%
  }
\end{table}

\begin{figure}[t]
    \centering
    \begin{minipage}[t]{0.54\linewidth}
        \centering
        \includegraphics[width=\linewidth]{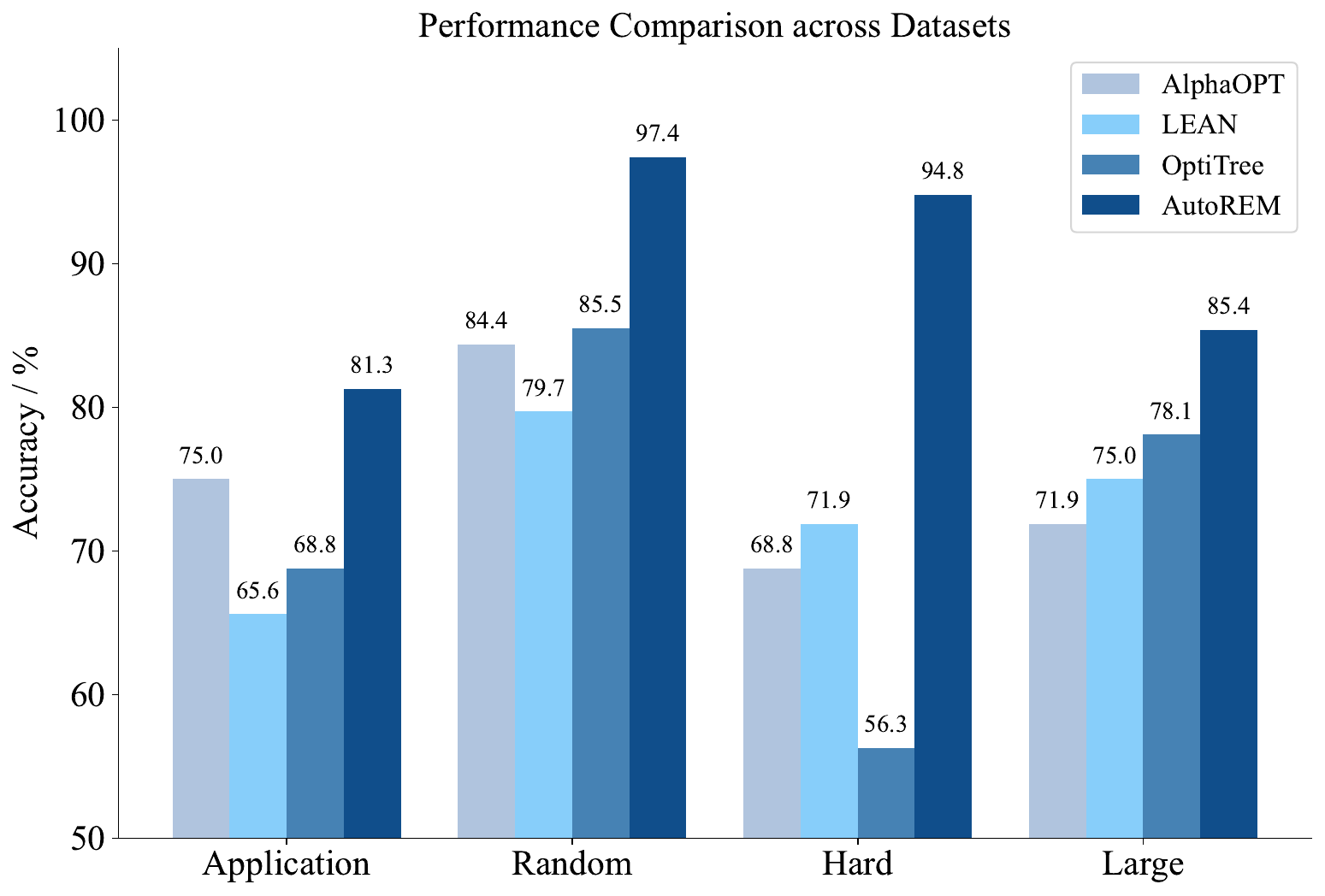}
        \caption{Comparison with LLM-based benchmarks.}
        \label{fig:apply}
    \end{minipage} 
    \hfill
    \begin{minipage}[t]{0.35\linewidth}
        \centering
        \includegraphics[width=\linewidth]{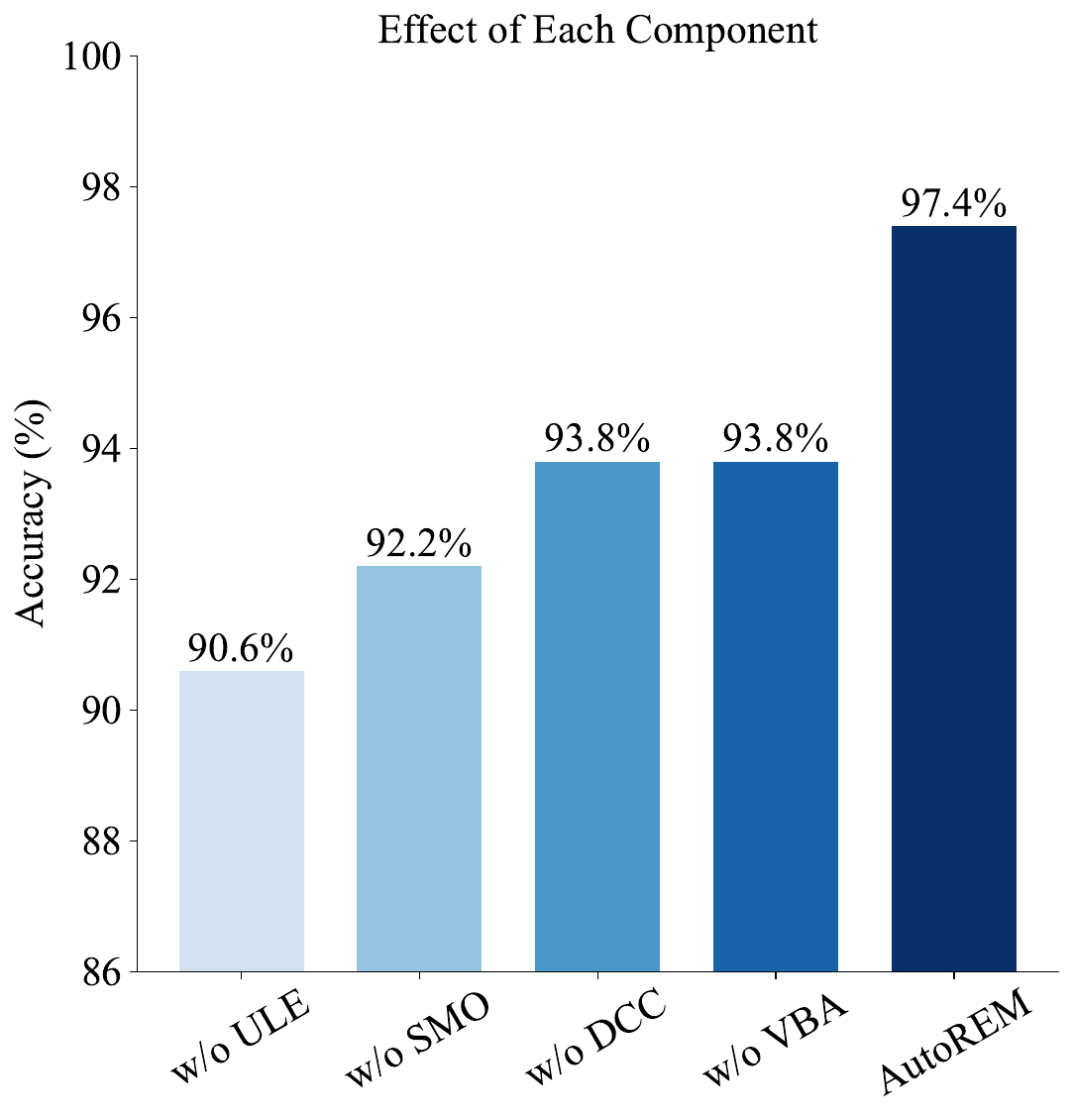}
        \caption{Ablation results.}
        \label{fig:abla}
    \end{minipage}
\end{figure}

\textbf{Application results.}
As shown in Figure~\ref{fig:apply}, on the end-to-end \emph{Application} dataset where all methods operate on natural-language problem descriptions, AutoREM achieves 81.3\% against the best baseline of 75.0\% (AlphaOPT), demonstrating strong practical capability for real-world RO problem solving.
When the same three formulation methods are evaluated on the pure reformulation datasets (bypassing any formulation step), the advantage of AutoREM becomes even more pronounced: on the \emph{Hard} dataset, the best baseline reaches only 71.9\% while AutoREM maintains 94.8\%, a gap exceeding 22\%.
This widening margin reveals that the bottleneck lies specifically in reformulation: these formulation-oriented methods lack the algebraic machinery for robust counterpart derivations, and it is precisely this capability that AutoREM's reformulation-specific memory supplies.

\begin{figure}[t]
    \centering
    \begin{minipage}[t]{0.46\linewidth}
        \centering
        \includegraphics[width=\linewidth]{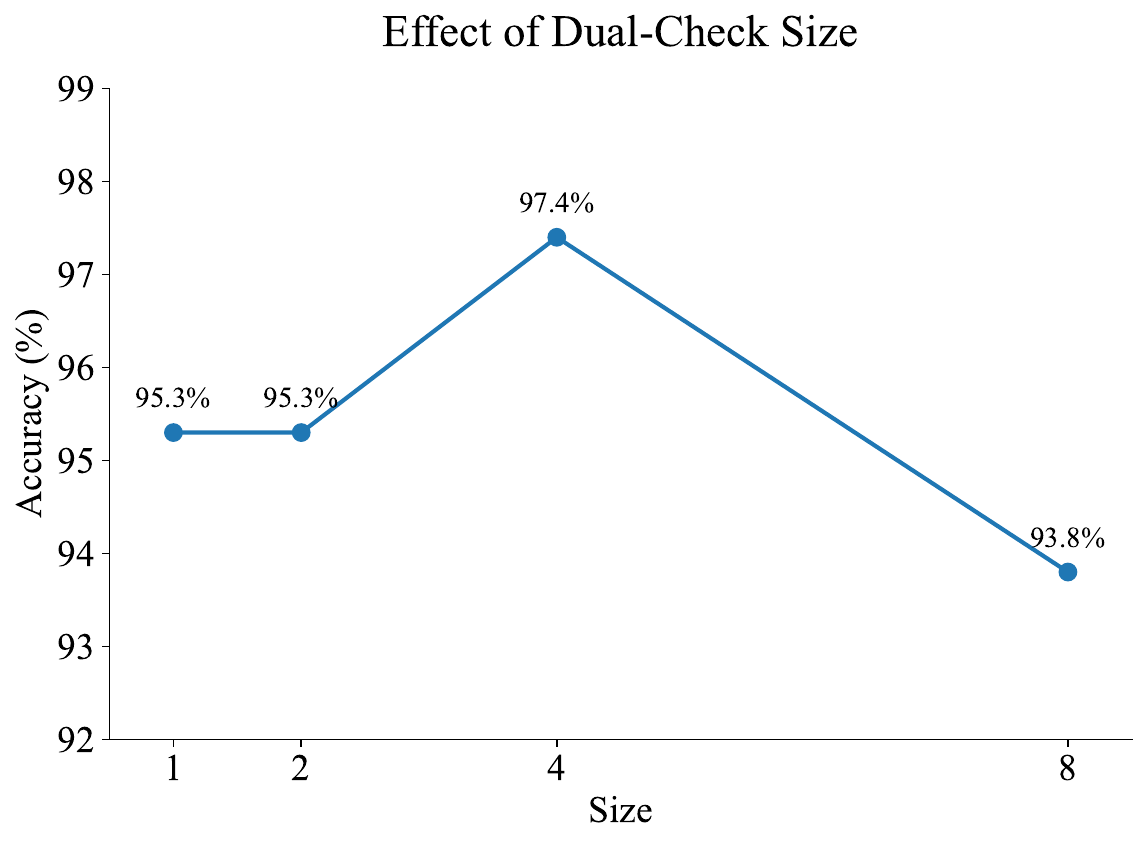}
        \caption{Effect of dual-check size $B$.}
        \label{fig:dc_size}
    \end{minipage}
    \hfill
    \begin{minipage}[t]{0.46\linewidth}
        \centering
        \includegraphics[width=\linewidth]{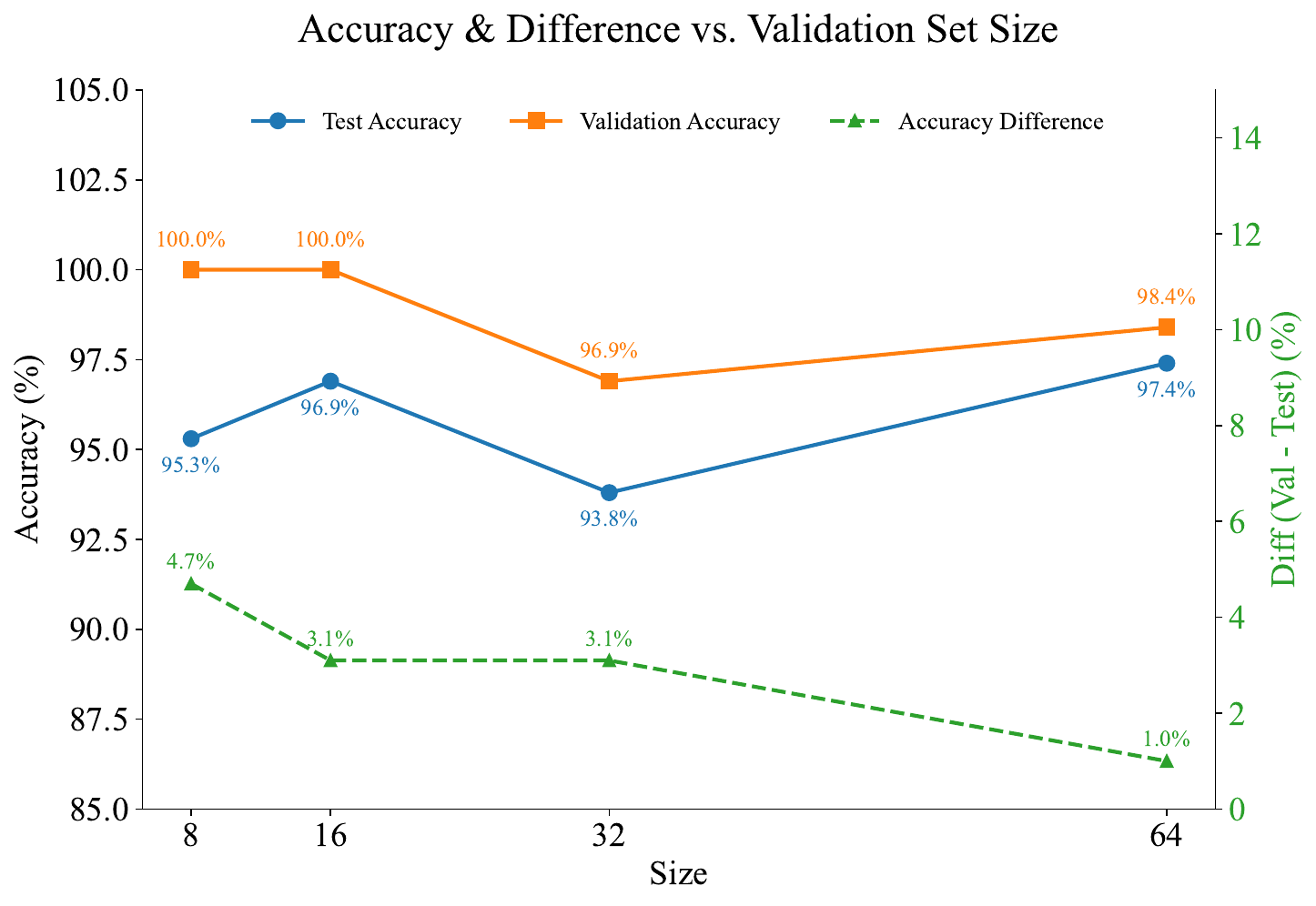}
        \caption{Effect of validation set size $|\mathcal{V}|$.}
        \label{fig:val_size}
    \end{minipage}
\end{figure}

\textbf{Ablation and hyperparameter analysis.}
Figure~\ref{fig:abla} shows that removing any components from ULE, SMO, DCC, and VBA can cause a non-trivial accuracy drop, demonstrating that each is indispensable.
Figures~\ref{fig:dc_size} and~\ref{fig:val_size} further show that the chosen defaults ($B=4$, $|\mathcal{V}|=64$) achieve the best accuracy, with performance remaining robust around these values.
Implementation details and a full sensitivity analysis are provided in Appendices~\ref{appx:ablation} and~\ref{appx:hyperparam}.

\section{Conclusion}

We introduced AutoRO-Bench and AutoREM, respectively the first benchmark and the first tuning-free memory-augmented framework for automated RO reformulation. AutoREM consistently achieves state-of-the-art accuracy across in-distribution and out-of-distribution settings, transfers across diverse base LLMs without re-adaptation, and outperforms LLM-based formulation methods on both the pure reformulation task and the end-to-end application task. One current limitation is that offline adaptation relies on solver-verified objective values as the supervision signal, which may not be available for all problem classes; future work could explore weaker supervision such as feasibility checks.
Extending the framework beyond LP-based robust counterparts to more complex RO is another important direction.
Further directions include conic and adjustable-robust models, stronger equivalence verification, and safeguards for untrusted inputs.

\begin{ack}
This research is supported by the National Research Foundation, Singapore and Infocomm Media Development Authority under its Trust Tech Funding Initiative, the National Natural Science Foundation of China, and the Ministry of Education, Singapore. Any opinions, findings and conclusions or recommendations expressed in this material are those of the author(s) and do not reflect the views of National Research Foundation, Singapore and Infocomm Media Development Authority.

Hanzhang Qin is supported by the Trust Tech Funding Initiative under grant DTC-IGC-07 (project: Multi-Party Evaluation: Blockchain-Enabled AI Safety and Verification Framework) and by the National Natural Science Foundation of China under grants U25A20450 and 62571374. Jinbiao Chen and Guanyi Wang are supported by the Ministry of Education, Singapore, under the Academic Research Fund (AcRF) Tier-1 grant (A-8000607-00-00) 22-5539-A0001 and Tier-2 grant T2EP20125-0030. The authors declare no competing interests.
\end{ack}
{
\small
\bibliographystyle{unsrtnat}
\bibliography{ref}
}

\clearpage
\appendix

\vbox{
\hrule height 4pt
\vskip 0.25in
\vskip -\parskip%
\centering
{\LARGE\bf Automated Reformulation of Robust Optimization via Memory-Augmented Large Language Models (Appendix)}
\vskip 0.29in
\vskip -\parskip
\hrule height 1pt
}

\section{Derivations of robust counterparts}
\label{app:rc-derivations}

This section derives the robust counterpart (RC) for each of the three
polyhedral uncertainty sets introduced in Section~\ref{sec:uncertainty-sets}.
All three derivations follow the same two-step template: (i) reformulate the
worst-case objective function or constraint  as an inner maximization problem, and (ii) replace this inner maximization with its LP dual, yielding a finite set of linear
rows.

Throughout, we work with the $i$-th uncertain constraint in the direct perturbation form
\[
  \sum_{j=1}^{n}\tilde{a}_{ij}\,x_{j} \leq b_{i},
  \qquad
  \tilde{a}_{ij} = \bar{a}_{ij} + \zeta_{ij},
\]
where $\boldsymbol{\zeta}_i\in\mathcal{U}_i$ is the perturbation vector, $\zeta_{ij}$ is its $j$-th component, and $\bar{a}_{ij}$ is the nominal coefficient.
Within each subsection the row index $i$ is fixed, so we write $\boldsymbol{\zeta}$ and $\zeta_j$ for $\boldsymbol{\zeta}_i$ and $\zeta_{ij}$ for brevity.
Requiring this to hold for every $\boldsymbol{\zeta}_i\in\mathcal{U}_i$ is equivalent to
\begin{equation}
\label{eq:app-worstcase}
  \sum_{j=1}^{n}\bar{a}_{ij}\,x_{j}
  + \max_{\boldsymbol{\zeta}_i\in\mathcal{U}_i}
      \sum_{j=1}^{n}\zeta_{ij}\,x_{j}
  \leq b_{i}.
\end{equation}
The inner maximization in~\eqref{eq:app-worstcase} is a linear program in
$\boldsymbol{\zeta}_i$ (for fixed $\mathbf{x}$); replacing it by its dual gives
the RC.

\subsection{Box uncertainty set}

For $\mathcal{U}_{\mathrm{box}}=\{\boldsymbol{\zeta}\mid|\zeta_{j}|\leq\delta_{j},\forall j\}$,
the inner problem is
\[
  \max_{\boldsymbol{\zeta}}\;\sum_{j}\zeta_{j}\,x_{j}
  \quad\text{s.t.}\quad |\zeta_{j}|\leq\delta_{j},\;\forall\,j.
\]
This separates over $j$: for each $j$, $\max_{|\zeta_j|\leq\delta_j}\zeta_j x_j = \delta_j|x_j|$, achieved at $\zeta_j^*=\delta_j\,\mathrm{sign}(x_j)$.
Introducing auxiliary variables $t_j\geq 0$ to linearize $\delta_j|x_j|$ (i.e., $t_j\geq\delta_j x_j$ and $t_j\geq -\delta_j x_j$) yields the LP robust counterpart:
\begin{equation}
\label{eq:app-box-rc}
\sum_{j=1}^{n}\bar{a}_{ij}\,x_{j} + \sum_{j=1}^{n}t_{j} \leq b_{i},
\qquad
t_j\geq\delta_j x_j,\quad t_j\geq -\delta_j x_j,\quad t_j\geq 0.
\end{equation}

\subsection{Budget uncertainty set}

For $\mathcal{U}_{\Gamma}=\bigl\{\boldsymbol{\zeta}\mid|\zeta_{j}|\leq\delta_{j},\forall j;\;
\sum_{j}|\zeta_{j}|/\delta_{j}\leq\Gamma_{i}\bigr\}$, the inner problem is
\[
  \max_{\boldsymbol{\zeta}}\;\sum_{j}\zeta_{j}\,x_{j}
  \quad\text{s.t.}\quad
  |\zeta_{j}|\leq\delta_{j},\;\forall\,j;\quad
  \sum_{j}\frac{|\zeta_{j}|}{\delta_{j}}\leq\Gamma_{i}.
\]
Substituting $u_{j}=\zeta_{j}/\delta_{j}$ (so $|u_{j}|\leq 1$ and $\sum_{j}|u_{j}|\leq\Gamma_{i}$)
and dualizing the budget constraint with multiplier $z_{i0}\geq 0$ and the
individual box constraints with multipliers $z_{ij}\geq 0$, strong duality gives
$z_{i0}+z_{ij}\geq\delta_j|x_j|$ for each $j$.
Splitting the absolute value into two linear constraints yields the LP robust counterpart:
\begin{equation}
\label{eq:app-budget-rc}
\sum_{j=1}^{n}\bar{a}_{ij}\,x_{j}
  + \Gamma_{i}\,z_{i0} + \sum_{j=1}^{n}z_{ij} \leq b_{i},
\quad
z_{i0}+z_{ij} \geq \pm\delta_{j}\,x_{j},\;\forall\,j,
\quad z_{ij}\geq 0,\;z_{i0}\geq 0.
\end{equation}
Setting $\Gamma_{i}=|\mathcal{S}_i|$ recovers the box RC~\eqref{eq:app-box-rc} (budget constraint becomes inactive).

\subsection{General polyhedral uncertainty set}

For $\mathcal{U}_{\mathrm{poly}}=\{\boldsymbol{\zeta}\mid\mathbf{F}\boldsymbol{\zeta}
\leq\mathbf{g}\}$, the uncertain row evaluates to
$\bar{\mathbf{a}}^{\!\top}\mathbf{x}+\boldsymbol{\zeta}^{\!\top}\mathbf{x}$,
so the inner problem is
\[
  \max_{\boldsymbol{\zeta}}\;\boldsymbol{\zeta}^{\!\top}\mathbf{x}
  \quad\text{s.t.}\quad \mathbf{F}\boldsymbol{\zeta}\leq\mathbf{g}.
\]
Associating dual variables $\boldsymbol{\lambda}\geq\mathbf{0}$ with the
constraints $\mathbf{F}\boldsymbol{\zeta}\leq\mathbf{g}$, the LP dual is
\[
  \min_{\boldsymbol{\lambda}\geq\mathbf{0}}\;
  \mathbf{g}^{\!\top}\boldsymbol{\lambda}
  \quad\text{s.t.}\quad
  \mathbf{F}^{\!\top}\boldsymbol{\lambda}=\mathbf{x}.
\]
By strong duality, the RC becomes
\begin{equation}
\label{eq:app-poly-rc}
\bar{\mathbf{a}}^{\!\top}\mathbf{x}
  + \mathbf{g}^{\!\top}\boldsymbol{\lambda} \leq b,
\quad
\mathbf{F}^{\!\top}\boldsymbol{\lambda} = \mathbf{x},
\quad
\boldsymbol{\lambda}\geq\mathbf{0}.
\end{equation}
This formulation subsumes the box and budget cases: both can be cast as
instances of $\mathcal{U}_{\mathrm{poly}}$ with appropriate choices of
$\mathbf{F}$ and $\mathbf{g}$.

\noindent\textbf{Extension to equality constraints.}
In practice (see Appendix~\ref{appx:datastep3}), the polyhedral uncertainty set may additionally include equality constraints $\mathbf{E}\boldsymbol{\zeta}=\mathbf{e}$, used to zero out coefficients outside the uncertain support $\mathcal{S}_i$.
Introducing free dual variables $\boldsymbol{\mu}$ for these constraints, the RC generalizes to
\begin{equation}
\label{eq:app-poly-rc-eq}
\bar{\mathbf{a}}^{\!\top}\mathbf{x}
  + \mathbf{g}^{\!\top}\boldsymbol{\lambda}
  + \mathbf{e}^{\!\top}\boldsymbol{\mu} \leq b,
\quad
\mathbf{F}^{\!\top}\boldsymbol{\lambda} + \mathbf{E}^{\!\top}\boldsymbol{\mu} = \mathbf{x},
\quad
\boldsymbol{\lambda}\geq\mathbf{0},\quad\boldsymbol{\mu}\ \text{free.}
\end{equation}

\section{RO reformulation task: data generation details}
\label{appx:data_gen}

This section supplements Section~4.1 with the full details of the data generation pipeline for the RO reformulation task. Algorithm~\ref{alg:data_gen} gives the complete generation procedure.

\begin{algorithm}[t]
\caption{Data generator for the RO reformulation task}
\label{alg:data_gen}
\begin{algorithmic}[1]
\Require $n$ (number of variables); $N$ (target instances); uncertainty type set $\mathcal{U}$; template set $\mathcal{T}$
\Ensure Dataset $\mathcal{D}$
\State $\mathcal{D} \leftarrow \emptyset$
\State $\mathrm{count} \leftarrow 0$
\While{$\mathrm{count} < N$}
  \State \textbf{Step 1: Nominal LP.} Sample objective coefficients $\mathbf{c}$ and direction; construct sparse feasible constraint polytope $(\mathbf{A}, \mathbf{b})$ with interior point $\mathbf{x}_0$ via the \textsc{FeasiblePolytope} algorithm (Appx.~\ref{appx:feasiblelp})
  \State \textbf{Step 2: Uncertainty types.} Assign an uncertainty type from $\mathcal{U}$ to each row
  \If{all rows deterministic} \textbf{continue} \EndIf
  \State \textbf{Step 3: Uncertainty parameters} (see Appx.~\ref{appx:datastep3})
  \For{each uncertain row $i$}
    \State Sample sparse support $\mathcal{S}_i \subseteq \{j : a_{ij} \neq 0\}$, $|\mathcal{S}_i| \ge 2$
    \If{Box} sample per-element deviations $\boldsymbol{\delta}_i$
    \ElsIf{Budget} sample $\boldsymbol{\delta}_i$ and budget $\Gamma_i$
    \ElsIf{Poly}
      \State $(\mathbf{F}, \mathbf{g}) \leftarrow \textsc{FeasiblePolytope}(|\mathcal{S}_i|,\, \boldsymbol{\zeta}^0,\, [-p_r, p_r],\, |\mathcal{S}_i|)$ \hfill (Appx.~\ref{appx:feasiblelp})
    \EndIf
  \EndFor
  \State \textbf{Step 4: Solve \& filter.} Solve via \texttt{RSOME}; discard if infeasible, unbounded, or degenerate
  \State \textbf{Step 5: Render.} Select $\mathcal{T}' \subseteq \mathcal{T}$; render \LaTeX\ using each $T_{b_0 b_1 b_2} \in \mathcal{T}'$; add records to $\mathcal{D}$
  \State $\mathrm{count} \leftarrow \mathrm{count} + 1$
\EndWhile
\State \Return $\mathcal{D}$
\end{algorithmic}
\end{algorithm}

\subsection{Problem structure and scale}
\label{appx:problem_structure}

\paragraph{Variable count and constraint count.}
Each RO instance is parameterized by a number of decision variables $n \in \{2, 3, 4, 5\}$.
The number of inequality/equality constraints equals $n$, i.e., the constraint matrix $\mathbf{A}\in\mathbb{R}^{n\times n}$.

\paragraph{Variable bounds.}
A nominal range $r \sim \mathrm{Uniform}(1, 10)$ (rounded to 1 decimal place) is sampled per instance.
A variable-bound type $\tau_x \in \{a,\, b,\, c\}$ is then selected uniformly at random, determining the box $[x_l, x_u]$ that every decision variable $x_i$ must satisfy:
\[
  \tau_x = a:\; x_l = -r,\; x_u = r;\qquad
  \tau_x = b:\; x_l = 0,\; x_u = r;\qquad
  \tau_x = c:\; x_l = -r,\; x_u = 0.
\]
The interior point $\mathbf{x}_0\in[x_l, x_u]^n$ used in Step~1 is sampled component-wise as $x_{0,i}\sim\mathrm{Uniform}(x_l, x_u)$, rounded to 1 decimal place.

\paragraph{Objective coefficients.}
The $n$ objective coefficients are sampled i.i.d.\ as $c_j\sim\mathrm{Uniform}(-10, 10)$, rounded to 1 decimal place.
The optimization direction (maximize or minimize) is chosen uniformly at random and is kept fixed for the entire instance.

\paragraph{Polyhedral perturbation range.}
The polyhedral range is set to $p_r = 0.1\cdot r$ (1 decimal place), where $r$ is the nominal range defined above.
A perturbation-bound type $\tau_p\in\{a,\,b,\,c\}$ is selected uniformly at random \emph{once per instance} and shared across all polyhedral uncertain rows:
\[
  \tau_p = a:\; p_l=-p_r,\; p_u=p_r;\qquad
  \tau_p = b:\; p_l=0,\; p_u=p_r;\qquad
  \tau_p = c:\; p_l=-p_r,\; p_u=0.
\]

\paragraph{Deterministic row guarantee.}
The generator enforces that \emph{exactly one} row (counting the objective as row~0 and the $n$ constraints as rows~1 through~$n$) is fully deterministic (uncertainty type~$= 0$).
If the nominal LP (Step~1) happens to produce an equality constraint at row $j$, that row is forced to be the deterministic row.
Otherwise, the deterministic row is chosen uniformly at random from $\{0, 1, \ldots, n\}$.
If the forced deterministic row is the objective (row~0), all $n$ constraint rows may be uncertain; otherwise, the objective row is assigned an uncertainty type drawn uniformly from $\{1, 2, 3\}$ (box / budget / polyhedral).
Each remaining constraint row is assigned an uncertainty type independently and uniformly from $\{1, 2, 3\}$.

\subsection{Step 4: solve and filter}
\label{appx:datastep4}

After the instance parameters are fully generated, the robust model is constructed and solved using \texttt{RSOME} (via \texttt{rsome.ro.Model}).
An instance is \emph{accepted} and added to the dataset only if \emph{all three} of the following conditions hold:
\begin{enumerate}
  \item \textbf{Not all-deterministic.} At least one row (objective or constraint) must have a positive uncertainty type; instances where every row is deterministic are discarded before solving.
  \item \textbf{Feasible and bounded.} The solver must return status \texttt{optimal}; instances that are infeasible, unbounded, or produce a solver error are discarded.
  \item \textbf{Non-degenerate solution.} The optimal solution $\mathbf{x}^*$ must not be identical to the variable lower or upper bound at every component, i.e., it is required that $\mathbf{x}^*$ is not coordinatewise equal to $\mathbf{x}_l$ or $\mathbf{x}_u$. Formally, the instance is rejected if $x^*_i \approx x_l$ or $x^*_i \approx x_u$ for \emph{all} $i$, as such solutions are considered degenerate boundary solutions that offer little reformulation challenge.
\end{enumerate}
Rejected instances are discarded entirely (all parameters are re-sampled), and the generator loops until the target number of valid instances is reached.

\subsection{Shared module: feasible polytope generator}
\label{appx:feasiblelp}

A key primitive used in two places of the generator is a procedure that constructs a random system of linear inequalities for which a prescribed point is guaranteed to be a strictly interior point. It is invoked as:
\begin{enumerate}
  \item \textbf{Nominal LP} (Step~1): generate the constraint matrix $\mathbf{A}$ and right-hand side $\mathbf{b}$ such that $\mathbf{x}_0$ is strictly feasible.
  \item \textbf{Polyhedral uncertainty set} (Step~3): generate the halfspace representation $(\mathbf{F}, \mathbf{g})$ of the perturbation polytope $\mathcal{U}_{\mathrm{poly}}$ such that the sampled interior point $\boldsymbol{\zeta}^0$ is guaranteed to lie strictly inside.
\end{enumerate}

\begin{algorithm}[t]
\caption{\textsc{FeasiblePolytope}: random feasible inequality system generator}
\label{alg:feasible_poly}
\begin{algorithmic}[1]
\Require Dimension $d$; interior point $\mathbf{v}^0 \in \mathbb{R}^d$; coefficient range $[l, u]$; number of rows $m$
\Ensure Matrix $\mathbf{A} \in \mathbb{R}^{m \times d}$, vector $\mathbf{b} \in \mathbb{R}^m$, directions $\mathbf{s} \in \{{\le}, {\ge}\}^m$
\For{$i = 1, \ldots, m$}
  \State Sample $\mathbf{a}_i \sim \mathrm{Uniform}([l,u]^d)$; clip $|a_{ij}| \ge 0.1$; round to 1 decimal
  \If{$d > 2$} apply random binary mask, keeping $\ge 2$ non-zero entries per row \EndIf
  \State Sample $s_i \sim \mathrm{Uniform}(\{{\le},\, {\ge}\})$
  \State $v_i \leftarrow \mathbf{a}_i^\top \mathbf{v}^0$
  \If{$s_i = {\le}$} $b_i \sim \mathrm{Uniform}(v_i,\; v_i + \|\mathbf{a}_i\|_1 \cdot |u|)$ \Comment{slack above $\mathbf{v}^0$}
  \Else\ $b_i \sim \mathrm{Uniform}(v_i - \|\mathbf{a}_i\|_1 \cdot |u|,\; v_i)$ \Comment{slack below $\mathbf{v}^0$}
  \EndIf
  \State Round $b_i$ to 2 decimal places
\EndFor
\State \Return $\mathbf{A}$, $\mathbf{b}$, $\mathbf{s}$
\end{algorithmic}
\end{algorithm}

\noindent\textbf{Equality constraints in the nominal LP.}
When constructing the nominal constraint system (invocation~1 above), a single row may additionally be converted to an equality: for $n \ge 3$, this occurs with probability $0.5$ for one randomly chosen row, setting $b_i = \mathbf{a}_i^\top \mathbf{x}_0$ exactly. Such a row is always designated as the deterministic row of the RO instance.

\subsection{Step 3: uncertainty parameter generation}
\label{appx:datastep3}

Both the nominal LP constraint matrix $\mathbf{A}$ and the polyhedral uncertainty polytopes are sparse by design. For $n \ge 3$, each row of $\mathbf{A}$ has at least two but generally fewer than $n$ non-zero entries, determined by the random binary mask in \textsc{FeasiblePolytope} (Algorithm~\ref{alg:feasible_poly}). Uncertainty is further sparsified at the row level: for each uncertain row $i$, a random subset $\mathcal{S}_i$ of $|\mathcal{S}_i| \ge 2$ indices is selected from the non-zero coefficient positions, so that only a portion of each row's coefficients is subject to perturbation. Uncertainty parameters are generated as follows.

\textbf{Box uncertainty.}
Per-element deviations are set as $\Delta_{i,j} = \max(0.1,\; \delta_{ij} \cdot |\bar{a}_{i,j}|)$ with $\delta_{ij} \sim \mathrm{Uniform}(0.05, 0.20)$, rounded to 1 decimal. The uncertainty set is:
\[
  \mathcal{U}_{\mathrm{box}} = \bigl\{\boldsymbol{\zeta}_i \in \mathbb{R}^{|\mathcal{S}_i|} : |\zeta_{i,j}| \le \Delta_{i,j},\; \forall j \in \mathcal{S}_i\bigr\}.
\]

\textbf{Budget uncertainty.}
The same per-element deviations $\Delta_{i,j}$ are used, with an additional budget $\Gamma_i \sim \mathrm{Uniform}(0, |\mathcal{S}_i|)$ (rounded to 1 decimal) limiting total perturbation:
\[
  \mathcal{U}_{\mathrm{bud}} = \Bigl\{\boldsymbol{\zeta}_i : |\zeta_{i,j}| \le \Delta_{i,j},\; \sum_{j \in \mathcal{S}_i} \frac{|\zeta_{i,j}|}{\Delta_{i,j}} \le \Gamma_i\Bigr\}.
\]

\textbf{Polyhedral uncertainty.}
A polyhedral range $p_r = 0.1r$ is set, and component-wise bounds $[p_l, p_u]$ are chosen uniformly from $\{[-p_r, p_r],\, [0, p_r],\, [-p_r, 0]\}$ (one choice shared across all polyhedral rows of the same instance). An interior point $\boldsymbol{\zeta}^0 \sim \mathrm{Uniform}([p_l, p_u]^{|\mathcal{S}_i|})$ is sampled. The halfspace representation is obtained by calling $\textsc{FeasiblePolytope}(|\mathcal{S}_i|,\, \boldsymbol{\zeta}^0,\, [-p_r, p_r],\, |\mathcal{S}_i|)$ and retaining only the first $|\mathcal{S}_i| - 1$ inequality rows to form $\mathbf{F} \in \mathbb{R}^{(|\mathcal{S}_i|-1)\times|\mathcal{S}_i|}$ and $\mathbf{g} \in \mathbb{R}^{|\mathcal{S}_i|-1}$. The resulting uncertainty set is:
\[
  \mathcal{U}_{\mathrm{poly}} = \bigl\{\boldsymbol{\zeta}_i \in \mathbb{R}^{|\mathcal{S}_i|} : \mathbf{F}\boldsymbol{\zeta}_i \le \mathbf{g},\; p_l \le \zeta_{i,j} \le p_u,\; \forall j \bigr\}.
\]
By construction, $\boldsymbol{\zeta}^0$ is strictly interior to $\mathcal{U}_{\mathrm{poly}}$, guaranteeing non-emptiness.

\subsection{Template dimensions}
\label{appx:template}

The template set $\mathcal{T}$ comprises a family of $8$ templates $T_{b_0 b_1 b_2}$, parameterized by three independent binary dimensions: explicit/implicit worst-case direction ($b_0$), vector/scalar notation style ($b_1$), and explicit/implicit uncertainty type annotation ($b_2$). Using diverse templates encourages AutoREM to learn notation-invariant reformulation knowledge, preventing the experience memory from simply memorizing surface-level text styles.

Each of the $8$ presentation templates $T_{b_0 b_1 b_2}$ is fully determined by three independent binary dimensions, summarized in Table~\ref{tab:template_dims}.

\begin{table}[t]
\centering
\small
\caption{The three binary dimensions parameterizing the $8$ \LaTeX\ presentation templates.}
\label{tab:template_dims}
\resizebox{\textwidth}{!}{
\begin{tabular}{@{}clll@{}}
\toprule
\textbf{Dim.} & \textbf{Name} & \textbf{Value $= 0$ (Implicit)} & \textbf{Value $= 1$ (Explicit)} \\
\midrule
$b_0$ & Worst-case direction
  & Universal quantifier $\forall\,\boldsymbol{\xi}\in\mathcal{U}$
  & Nested operator $\min_{\boldsymbol{\xi}\in\mathcal{U}}$ / $\max_{\boldsymbol{\xi}\in\mathcal{U}}$ \\[4pt]
$b_1$ & Notation style
  & Scalar: each term written element-wise
  & Vector/matrix: compact inner-product form \\[4pt]
$b_2$ & Uncertainty type annotation
  & No label; set given by math definition only
  & Named label, e.g., ``Box Uncertainty'' \\
\bottomrule
\end{tabular}
}
\end{table}

\subsection{Hard dataset configuration}
\label{appx:hard_dataset}

The \emph{Hard} dataset is generated by the same pipeline as the \emph{Random} dataset (Algorithm~\ref{alg:data_gen}), but with the random sampling constrained so that each instance simultaneously satisfies four structural conditions identified as failure modes of the base LLM during preliminary experiments:
\begin{itemize}
  \item all uncertain constraints use polyhedral uncertainty;
  \item the objective direction is maximization;
  \item the uncertain constraints include both $\geq$ and $\leq$ directions;
  \item decision variables are allowed to take both positive and negative values.
\end{itemize}
The decision-variable count remains $n \in \{2, 3, 4, 5\}$ as in the Random split, so the accuracy gap relative to Random is attributable to these structural difficulty factors rather than to dimensionality.

\subsection{Split independence and input length}
\label{appx:evaluation_protocol}

The original reformulation test splits contain 64 \emph{Random}, 32 \emph{Hard}, and 32 \emph{Large} instances. Training, validation, and these test instances are independently generated: no underlying problem instance, uncertainty-parameter set, or rendered variant is reused across them. The generator and presentation-template family are shared, rather than the generated instances. The original 32-instance \emph{Application} set is evaluated with the frozen memory learned from the reformulation training distribution.


Table~\ref{tab:input_token_statistics} characterizes the problem descriptions using the DeepSeek tokenizer. Only each instance's description field is encoded. System prompts, memory, and generated outputs are excluded. \emph{Large} has 2.19 times the mean description length of \emph{Random}. The generator increases both the number of variables and the number of nominal constraints from $n\in\{2,3,4,5\}$ to $n\in\{6,7\}$, with the constraint count set to $n$. Thus model dimension, uncertainty descriptions, numerical matrices, and context length increase together.

\begin{table}[t]
\centering
\caption{Input-description token statistics for the original reformulation test splits.}
\label{tab:input_token_statistics}
\begin{tabular}{lrrr}
\toprule
Split & Instances & Mean tokens & Token range \\
\midrule
Random & 64 & 720.00 & 282--1,450 \\
Hard & 32 & 885.97 & 348--1,812 \\
Large & 32 & 1,579.66 & 882--2,419 \\
\bottomrule
\end{tabular}
\end{table}

\section{RO application dataset: literature-motivated parameter generation}
\label{appx:poly_types}

\paragraph{Source-problem selection and coverage.}
The original Application set is a curated stress test rather than an exhaustive sample of industrial robust optimization. Its 16 nominal problems are drawn equally from IndustryOR and OptiBench, with eight problems from each source and two robust variants per problem. Selection prioritizes meaningful uncertainty in multiple rows: the relevant coefficients have physical or economic interpretations that can plausibly vary. Many transportation, assignment, and allocation formulations in the source benchmarks predominantly use structural coefficients such as $0$, $1$, or $-1$. Imposing continuous uncertainty on these coefficients can be physically artificial. Fifteen selected nominal problems support multi-row uncertainty, while one has only objective uncertainty. For each nominal problem, the first variant uses the reformulation benchmark's numerical parameter-generation procedure, while the second uses the literature-motivated settings detailed below. This gives diverse and interpretable tests within the supported uncertainty families, while retaining the limitations of a small curated set. The full problem-class, variable-count, and uncertain-row statistics are reported in Appendix~\ref{appx:app_stats}.

For the second RO instance of each nominal problem in the RO application dataset, uncertainty parameters are generated following practically-motivated settings from the RO literature.

\textbf{Uncertain row selection.}
A row is made uncertain only if it contains more than two coefficients that are simultaneously non-zero and not equal to $\pm 1$. Let $\mathcal{S}$ denote the index set of such qualifying coefficients, $m = |\mathcal{S}|$ the number of uncertain coefficients, and define the scale baseline $s = \frac{1}{m}\sum_{j \in \mathcal{S}}|\bar{a}_{j}|$, where $\bar{\mathbf{a}}$ are the nominal coefficients of the row. Note that $m$ is distinct from $n$ (the number of decision variables defined in Section~\ref{sec:uncertainty-sets}).

\textbf{Box uncertainty~\citep{ben-tal2000robust}.}
Each deviation is $\Delta_j = p \cdot |\bar{a}_j|$ for $j \in \mathcal{S}$, where $p$ is randomly chosen from $\{0.01\%,\, 0.1\%,\, 1\%\}$:
\[
  \mathcal{U}_{\mathrm{box}} = \{\boldsymbol{\xi} : |\xi_j| \le \Delta_j,\ \forall j \in \mathcal{S}\}.
\]

\textbf{Budget uncertainty.}
The same deviation $\Delta_j$ as for box is used. To reflect varying degrees of conservatism, the budget parameter $\Gamma_b$ is randomly chosen to be $40\%$, $60\%$, or $80\%$ of $m$. Note that $\Gamma_b$ here is the budget threshold and is distinct from the robustness level parameter $\Gamma$ used in the polyhedral sets below:
\[
  \mathcal{U}_{\mathrm{bud}} = \Bigl\{\boldsymbol{\xi} : |\xi_j| \le \Delta_j,\; \sum_{j \in \mathcal{S}} \tfrac{|\xi_j|}{\Delta_j} \le \Gamma_b\Bigr\}.
\]

\textbf{Polyhedral uncertainty~\citep{bandi2012tractable}.}
One of the following four statistically-motivated types is randomly selected. All use a fixed robustness level $\Gamma = 2.0$. Unlike box and budget uncertainty where all parameters are presented as explicit numerical values, these sets define their constraint bounds through symbolic expressions involving $m$ and sampled distribution parameters (e.g., $\Gamma\sigma\sqrt{m}$, $m/\lambda$). These expressions are retained symbolically in the problem description and must be preserved accordingly in the reformulation output to avoid numerical precision loss.

\textit{CLT-based.} Derived from the Central Limit Theorem for i.i.d.\ random variables with mean $\mu$ and standard deviation $\sigma$:
\[
  \mathcal{U}_{CLT} = \Bigl\{\boldsymbol{\xi} \Bigm| -\Gamma\sigma\sqrt{m} \le \textstyle\sum_{j \in \mathcal{S}}\xi_j - m\mu \le \Gamma\sigma\sqrt{m}\Bigr\},
\]
where $\mu \sim \mathrm{Uniform}(-0.1s,\, 0.1s)$ and $\sigma \sim \mathrm{Uniform}(0.05s,\, 0.2s)$.

\textit{Heavy-tail.} For heavy-tailed distributions with tail index $\alpha \in (1,2]$, sums scale as $m^{1/\alpha}$ rather than $\sqrt{m}$:
\[
  \mathcal{U}_{HT} = \Bigl\{\boldsymbol{\xi} \Bigm| {-\Gamma} \le \frac{\sum_{j \in \mathcal{S}}\xi_j - m\mu}{m^{1/\alpha}} \le \Gamma\Bigr\},
\]
where $\mu \sim \mathrm{Uniform}(-0.1s,\, 0.1s)$ and $\alpha = 1.5$.

\textit{Typical set -- exponential.} Based on the typical set of i.i.d.\ Exponential($\lambda$) variables, with non-negative perturbations:
\[
  \mathcal{U}^{E} = \Bigl\{\boldsymbol{\xi} \Bigm| \tfrac{m}{\lambda} - \tfrac{\sqrt{m}}{\lambda}\Gamma \le \textstyle\sum_{j \in \mathcal{S}}\xi_j \le \tfrac{m}{\lambda} + \tfrac{\sqrt{m}}{\lambda}\Gamma,\ \xi_j \ge 0\Bigr\},
\]
where $\lambda \sim \mathrm{Uniform}\!\left(\tfrac{1}{0.2s},\, \tfrac{1}{0.05s}\right)$.

\textit{Typical set -- uniform.} Based on the typical set of i.i.d.\ Uniform($a$, $b$) variables:
\[
  \mathcal{U}^{U} = \Bigl\{\boldsymbol{\xi} \Bigm| m\cdot\tfrac{a+b}{2} - \Gamma\sqrt{m} \le \textstyle\sum_{j \in \mathcal{S}}\xi_j \le m\cdot\tfrac{a+b}{2} + \Gamma\sqrt{m},\ a \le \xi_j \le b\Bigr\},
\]
where $a,b \sim \mathrm{Uniform}(-0.2s,\, 0.2s)$ with $a < b$.

\section{RO application dataset: problem description format}
\label{appx:app_format}

Each instance in the RO application dataset consists of the original natural-language problem description from IndustryOR or OptiBench, followed by an appended \textbf{Robust Extension} paragraph. The extension begins with a preamble explaining the robust optimization setup (uncertain parameters modeled as nominal plus perturbation, worst-case feasibility requirement), lists each uncertain row with its uncertainty set description, and closes with a statement of the problem objective.

\noindent\textbf{Format of the Robust Extension.} The appended text follows this structure:

\begin{enumerate}
  \item \textbf{Preamble.} A fixed paragraph explaining that certain parameters are uncertain, perturbations are confined to uncertainty sets, and the goal is a worst-case feasible decision.
  \item \textbf{Uncertain parameter descriptions.} One bullet per uncertain row, identifying the physical meaning of the uncertain coefficients and specifying the uncertainty set. The set is described as:
    \begin{itemize}
      \item \textit{Box:} ``Each coefficient may deviate by at most $\pm[\Delta_1, \ldots, \Delta_m]$ from its nominal value.''
      \item \textit{Budget:} Same as box, followed by ``the sum of normalized deviations ($\sum|\xi_j|/\Delta_j$) across all $m$ uncertain coefficients must not exceed $\Gamma_b = \langle\text{value}\rangle$.''
      \item \textit{Polyhedral:} ``The perturbations $\boldsymbol{\xi}$ are confined to a polyhedral uncertainty set with each $\xi_i$ bounded within $[l, u]$, and satisfying the following $k$ linear constraint(s): $\langle\text{constraint list}\rangle$.''
    \end{itemize}
  \item \textbf{Closing statement.} ``The robust model seeks a decision that guarantees [maximizing/minimizing] the objective value under all worst-case realizations within the uncertainty sets above.''
\end{enumerate}

\noindent\textbf{Example.} The following is an excerpt from one instance (a bakery production problem from OptiBench), illustrating the box uncertainty format:

\begin{quote}
\textit{Robust Extension:}
In this robust version, certain parameters in the problem are treated as uncertain \ldots

The following parameters are subject to uncertainty:

\textbullet\ \textit{Objective coefficients}: The profit per cake type may vary due to ingredient cost and pricing fluctuations. Each relevant coefficient may fluctuate independently. The maximum absolute deviation for each coefficient is $\pm[1.97,\, 1.64,\, 2.46]$.

\textbullet\ \textit{Batter required per cake}: The batter quantity per cake may vary due to recipe variation. Each relevant coefficient may fluctuate independently. The maximum absolute deviation for each coefficient is $\pm[71.76,\, 89.70,\, 80.73]$.

\textbullet\ \textit{Milk required per cake}: May vary. The maximum absolute deviation for each coefficient is $\pm[26.67,\, 40.00,\, 46.67]$.

The robust model seeks a decision that guarantees maximizing the objective value under all worst-case realizations within the uncertainty sets above.
\end{quote}

\section{RO application dataset: statistics}
\label{appx:app_stats}

This section reports detailed statistics of the RO application dataset and contrasts them with the RO reformulation dataset (Random split). Since each application instance is curated from a real-world problem rather than algorithmically generated, its problem-level characteristics differ substantively from the reformulation dataset, while the per-instance reformulation challenge---measured by the number of uncertain rows---remains broadly comparable. We index the 16 nominal problems as P1--P16 (P1--P8 from IndustryOR, P9--P16 from OptiBench), each contributing 2 instances as described in Section~4.1.

\paragraph{Problem class.}
The 16 nominal problems comprise 7 LPs and 9 MILPs, yielding 14 LP and 18 MILP instances after the two-instance expansion. The reformulation dataset is exclusively LP, since its automatic generator (Algorithm~\ref{alg:data_gen}) does not produce integer variables. The presence of MILPs in the application dataset stress-tests AutoREM's ability to apply LP-derived reformulation memory to mixed-integer settings.

\paragraph{Decision variables.}
Application instances contain 2 to 13 decision variables (median 3), spanning a wider range than the reformulation Random split's $\{2, 3, 4, 5\}$. The 13-variable instance (P1) substantially exceeds the dimensionality of any reformulation training instance, providing an additional dimensionality stress test.

\paragraph{Uncertain rows.}
The number of uncertain rows, i.e., the count of objective and constraint rows containing uncertain coefficients, is the most direct indicator of reformulation difficulty. Each additional uncertain row requires an independent counterpart derivation, increasing both reasoning depth and output context length. Application instances contain 1 to 4 uncertain rows, with 75\% (24/32) containing 3 or 4. By construction, every reformulation instance with $n$ decision variables contains exactly $n$ uncertain rows (one of the $n{+}1$ rows is forced deterministic; see Appendix~\ref{appx:problem_structure}), giving 2--5 uncertain rows for the Random split. Thus, despite the application dataset's smaller size, its per-instance reformulation challenge is broadly comparable to that of the reformulation dataset.

\begin{table}[t]
\centering
\caption{Detailed statistics of the RO application dataset (32 instances). Each source problem produces 2 instances.}
\label{tab:app_stats}
\small
\begin{tabular}{llcl}
\toprule
\textbf{Attribute} & \textbf{Value} & \textbf{\# Instances} & \textbf{Source Problems} \\
\midrule
\multirow{2}{*}{Problem class}
  & LP   & 14 & P1, P2, P6, P8, P9, P11, P13 \\
  & MILP & 18 & P3, P4, P5, P7, P10, P12, P14, P15, P16 \\
\midrule
\multirow{6}{*}{\# decision variables}
  & 2  & 12 & P4, P8, P13, P14, P15, P16 \\
  & 3  & 8  & P3, P5, P7, P10 \\
  & 4  & 4  & P9, P12 \\
  & 5  & 4  & P6, P11 \\
  & 6  & 2  & P2 \\
  & 13 & 2  & P1 \\
\midrule
\multirow{4}{*}{\# uncertain rows}
  & 1 & 2  & P3 (objective only) \\
  & 2 & 6  & P8, P14, P16 \\
  & 3 & 12 & P1, P7, P10, P12, P13, P15 \\
  & 4 & 12 & P2, P4, P5, P6, P9, P11 \\
\bottomrule
\end{tabular}
\end{table}

\section{Sample dataset instances}
\label{appx:examples}

This section presents one concrete instance from each dataset to complement the abstract descriptions in Section~4.1. We use the last instance of each dataset.

\subsection{RO reformulation dataset instance (Random split)}

The instance has identifier \texttt{5\_16\_T011}, with $n=5$ decision variables and $5$ constraints. The verified ground-truth solution is $\mathbf{x}^* = [1.8,\; 1.0231,\; 1.8,\; 0,\; 0.8]^\top$ with optimal value $f^* = 29.6985$.

\paragraph{Problem (LLM input).}
\allowdisplaybreaks
{\footnotesize
\begin{align*}
\text{Maximize} \quad & \mathbf{c}^\top \mathbf{x} && \forall \mathbf{c} \in \mathcal{U}_{c,Bud} && \text{(Budgeted Uncertainty Objective)} \\
\text{subject to} \quad & \mathbf{a}_{1}^\top \mathbf{x} \ge b_{1} && \forall \mathbf{a}_{1} \in \mathcal{U}_{a,Box} && \text{(Box Uncertainty Constraint)} \\
& \mathbf{a}_{2}^\top \mathbf{x} \le b_{2} && && \text{(Deterministic Constraint)} \\
& (\mathbf{a}_{3}^{nom} + \boldsymbol{\delta}_{a3})^\top \mathbf{x} \ge b_{3} && \forall \boldsymbol{\delta}_{a3} \in \mathcal{U}_{a3,P} && \text{(Polyhedral Uncertainty Constraint)} \\
& (\mathbf{a}_{4}^{nom} + \boldsymbol{\delta}_{a4})^\top \mathbf{x} \le b_{4} && \forall \boldsymbol{\delta}_{a4} \in \mathcal{U}_{a4,P} && \text{(Polyhedral Uncertainty Constraint)} \\
& (\mathbf{a}_{5}^{nom} + \boldsymbol{\delta}_{a5})^\top \mathbf{x} \ge b_{5} && \forall \boldsymbol{\delta}_{a5} \in \mathcal{U}_{a5,P} && \text{(Polyhedral Uncertainty Constraint)}
\end{align*}
\begin{align*}
\text{where:} \quad
& \mathcal{U}_{c,Bud} = \{ \mathbf{c} \in \mathbb{R}^{5} : \mathbf{c} = \mathbf{c}^{nom} + \mathbf{D}_c \boldsymbol{\xi}_c, \|\boldsymbol{\xi}_c\|_{\infty} \le 1, \|\boldsymbol{\xi}_c\|_1 \le \Gamma_c \} \\
& \mathcal{U}_{a,Box} = \{ \mathbf{a}_{1} \in \mathbb{R}^{5} : \mathbf{a}_{1} = \mathbf{a}_{1}^{nom} + \mathbf{D}_{a1} \boldsymbol{\xi}_{a1}, \|\boldsymbol{\xi}_{a1}\|_{\infty} \le 1 \} \\
& \mathcal{U}_{a3,P} = \{ \boldsymbol{\delta}_{a3} \in \mathbb{R}^{5} : \mathbf{M}_{a3} \boldsymbol{\delta}_{a3} \le \mathbf{q}_{a3}, \quad \mathbf{E}_{a3} \boldsymbol{\delta}_{a3} = \mathbf{e}_{a3} \} \\
& \mathcal{U}_{a4,P} = \{ \boldsymbol{\delta}_{a4} \in \mathbb{R}^{5} : \mathbf{M}_{a4} \boldsymbol{\delta}_{a4} \le \mathbf{q}_{a4}, \quad \mathbf{E}_{a4} \boldsymbol{\delta}_{a4} = \mathbf{e}_{a4} \} \\
& \mathcal{U}_{a5,P} = \{ \boldsymbol{\delta}_{a5} \in \mathbb{R}^{5} : \mathbf{M}_{a5} \boldsymbol{\delta}_{a5} \le \mathbf{q}_{a5}, \quad \mathbf{E}_{a5} \boldsymbol{\delta}_{a5} = \mathbf{e}_{a5} \} \\
& \mathbf{x} = [x_{1}, x_{2}, x_{3}, x_{4}, x_{5}]^\top \\
& \mathbf{c}^{nom} = [7.7, -1.8, 8.8, -2.9, 2.7]^\top \\
& \mathbf{a}_{1}^{nom} = [0, 0, -0.1, 0, -0.7]^\top \\
& b_{1} = -1 \\
& \mathbf{a}_{2} = [-0.2, -1.3, 0, -0.7, -1.3]^\top \\
& b_{2} = -2.73 \\
& \mathbf{a}_{3}^{nom} = [-0.4, 0, 0, 0, -1.1]^\top \\
& b_{3} = -2.64 \\
& \mathbf{a}_{4}^{nom} = [0, 1.8, -1.4, -0.2, 0.9]^\top \\
& b_{4} = 6.8 \\
& \mathbf{a}_{5}^{nom} = [0.3, 1, -1.1, 0, 0]^\top \\
& b_{5} = -2.76 \\
& \mathbf{D}_c = \text{diag}(0, 0, 0, 0.5, 0.5) \\
& \Gamma_c = 0.8 \\
& \mathbf{D}_{a1} = \text{diag}(0, 0, 0.1, 0, 0.1) \\
& \mathbf{M}_{a3} = \begin{bmatrix} -0.2 & 0 & 0 & 0 & -0.2 \\ 1 & 0 & 0 & 0 & 0 \\ -1 & 0 & 0 & 0 & 0 \\ 0 & 0 & 0 & 0 & 1 \\ 0 & 0 & 0 & 0 & -1 \end{bmatrix} \\
& \mathbf{q}_{a3} = [0.06, 0, 0.2, 0, 0.2]^\top \\
& \mathbf{E}_{a3} = \begin{bmatrix} 0 & 1 & 0 & 0 & 0 \\ 0 & 0 & 1 & 0 & 0 \\ 0 & 0 & 0 & 1 & 0 \end{bmatrix} \\
& \mathbf{e}_{a3} = [0, 0, 0]^\top \\
& \mathbf{M}_{a4} = \begin{bmatrix} 0 & -0.2 & 0 & 0 & -0.1 \\ 0 & -0.1 & 0 & 0.1 & 0 \\ 0 & 1 & 0 & 0 & 0 \\ 0 & -1 & 0 & 0 & 0 \\ 0 & 0 & 0 & 1 & 0 \\ 0 & 0 & 0 & -1 & 0 \\ 0 & 0 & 0 & 0 & 1 \\ 0 & 0 & 0 & 0 & -1 \end{bmatrix} \\
& \mathbf{q}_{a4} = [0.04, 0.01, 0, 0.2, 0, 0.2, 0, 0.2]^\top \\
& \mathbf{E}_{a4} = \begin{bmatrix} 1 & 0 & 0 & 0 & 0 \\ 0 & 0 & 1 & 0 & 0 \end{bmatrix} \\
& \mathbf{e}_{a4} = [0, 0]^\top \\
& \mathbf{M}_{a5} = \begin{bmatrix} -0.1 & -0.1 & 0 & 0 & 0 \\ 1 & 0 & 0 & 0 & 0 \\ -1 & 0 & 0 & 0 & 0 \\ 0 & 1 & 0 & 0 & 0 \\ 0 & -1 & 0 & 0 & 0 \end{bmatrix} \\
& \mathbf{q}_{a5} = [0.04, 0, 0.2, 0, 0.2]^\top \\
& \mathbf{E}_{a5} = \begin{bmatrix} 0 & 0 & 1 & 0 & 0 \\ 0 & 0 & 0 & 1 & 0 \\ 0 & 0 & 0 & 0 & 1 \end{bmatrix} \\
& \mathbf{e}_{a5} = [0, 0, 0]^\top \\
& 0 \le x_j \le 1.8, \quad \forall j \in \{1, \dots, 5\} \qquad \text{(Variable Bounds)}
\end{align*}
}

\subsection{RO application dataset instance}

The instance has identifier $32$, sourced from OptiBench~\citep{yang2025optibench}. The verified ground-truth optimal value is $f^* = 1761.568$.

\paragraph{Problem (LLM input).}
\begin{quote}
\small
There has been an oil spill in the ocean and ducks need to be taken to shore to be cleaned either by boat or by canoe. A boat can take 10 ducks per trip while a canoe can take 8 ducks per trip. Since the boats are motor powered, they take 20 minutes per trip while the canoes take 40 minutes per trip. In order to avoid further environmental damage, there can be at most 12 boat trips and at least 60\% of the trips should be by canoe. If at least 300 ducks need to be taken to shore, how many of each transportation method should be used to minimize the total amount of time needed to transport the ducks?

Robust Extension:

In this robust version, certain parameters in the problem are treated as uncertain --- they are not fixed at their nominal values. Instead, each uncertain parameter is modeled as its nominal value plus a perturbation, and it is this perturbation that is confined within specified uncertainty sets. The goal is to find a decision plan that remains feasible and near-optimal under all possible realizations of these uncertain parameters (the worst-case guarantee).

The following parameters are subject to uncertainty:

\textbullet\ Objective coefficients: The time per trip for boats and canoes may vary due to water conditions. Each coefficient may deviate by at most $0.10\%$ of its nominal value. Additionally, the combined perturbation is limited: the sum of normalized deviations $(\sum |\xi_j|/d_j)$ must not exceed $\Gamma = 1.2$, which means on average at most $60\%$ of the uncertain coefficients can simultaneously reach their maximum deviation. This captures the realistic situation where large simultaneous deviations across all parameters are unlikely.

\textbullet\ The number of ducks transported per trip may vary due to animal behavior. This uncertainty set accounts for heavy-tailed distributions (tail index $\alpha = 1.5$): the normalized sum $(\Sigma \xi_i - n\mu) / n^{1/\alpha}$ must lie within $[-\Gamma, +\Gamma]$, so the sum range is $[n\mu - \Gamma \cdot n^{1/\alpha}, n\mu + \Gamma \cdot n^{1/\alpha}]$. Compared to the CLT-based set, the normalization factor $n^{1/\alpha}$ grows more quickly than $n^{1/2}$, allowing for heavier tails. Each component is bounded within $[\mu - \Gamma \cdot n^{1/\alpha-1}, \mu + \Gamma \cdot n^{1/\alpha-1}]$ (parameters: $\mu = -0.8139$, $\alpha = 1.5$, $\Gamma = 2.0$, $n = 2$).

\textbullet\ \textit{Variable type}: The decision variables ($x_1, x_2$) represent trips by each transport method. Since these quantities are inherently discrete, all decision variables are constrained to be non-negative integers ($x_i \in \mathbb{Z}_+$).

The robust model seeks a decision that guarantees minimizing the objective value under all worst-case realizations within the uncertainty sets above.
\end{quote}

\section{Evaluation reliability: mathematical and coder audits}
\label{appx:extended_evaluation}

A matching objective is an automatic evaluation signal, not a certificate that the generated counterpart is mathematically equivalent to the input robust model. In particular, an incorrect model can share an optimum with the reference model on a particular instance. Comparing arbitrary formulations is also complicated by equivalent representations involving permutations, scaling, redundant constraints, auxiliary variables, and extended formulations~\citep{xing2024humanaligned,zhai2025equivamap}. The model and coder audits provide additional evidence for the evaluation reliability of objective matching.

\subsection{Expert audit of mathematical counterparts}
\label{appx:structural_audit}

Three operations research experts manually audited all 128 outputs from one pre-selected \emph{Random} run: 64 Base LLM outputs and 64 AutoREM outputs. They first judged each mathematical counterpart without viewing its generated code, and then examined whether the coder faithfully implemented the counterpart. This separates errors in the reformulator's mathematical output from changes introduced during coding. Table~\ref{tab:structural_audit} shows model-level judgments and the original objective-matching outcomes. ``Model correct'' refers to the reformulator's mathematical counterpart, before coding. The Base LLM produced 8 incorrect mathematical counterparts, whereas AutoREM produced 1. Three of the Base LLM's 58 objective-positive outputs were structurally incorrect, compared with none of AutoREM's 62 objective-positive outputs. AutoREM's improvement is therefore 10.9\% under direct model judgment (56/64 versus 63/64), compared with 6.3\% under objective matching (58/64 versus 62/64). The imperfect objective metric does not account for the observed advantage.

\begin{table}[t]
\centering
\small
\setlength{\tabcolsep}{3.5pt}
\caption{Structural audit of 64 outputs per method. ``Objective only'' means that the solved objective matches the reference despite an incorrect mathematical counterpart. ``Model only'' means a correct mathematical counterpart whose evaluated objective does not match.}
\label{tab:structural_audit}
\begin{tabular}{lrrrrrr}
\toprule
Method & \shortstack{Both\\correct} & \shortstack{Objective\\only} & \shortstack{Model\\only} & \shortstack{Both\\wrong} & \shortstack{Objective\\accuracy} & \shortstack{Model\\accuracy} \\
\midrule
Base LLM & 55 & 3 & 1 & 5 & 90.6\% & 87.5\% \\
AutoREM & 62 & 0 & 1 & 1 & 96.9\% & 98.4\% \\
\bottomrule
\end{tabular}
\end{table}

\subsection{Coder fidelity and error attribution}
\label{appx:coder_audit}

Table~\ref{tab:coder_audit} shows coder fidelity on the same 128 outputs. Among the 56 correct Base LLM counterparts, the coder introduced one structural error (1/56, or 1.8\%), which changed the objective. Among the 63 correct AutoREM counterparts, it introduced two structural errors (2/63, or 3.2\%). Only one changed the objective (1/63, or 1.6\%), while the other was nonbinding. No incorrect mathematical counterpart was verified as repaired by the coder. The Base LLM's six objective failures comprise five reformulator-caused failures and one coder-only failure; AutoREM's two comprise one of each. Correcting only the confirmed coder-only errors changes objective accuracy from 58/64 to 59/64 for Base LLM and from 62/64 to 63/64 for AutoREM. The gap remains 4/64, or 6.25\%. The main improvement on this run consequently comes from reducing incorrect reformulations from eight to one, with small and balanced outcome-affecting coder errors.

\begin{table}[t]
\centering
\small
\caption{Coder audit and objective-failure attribution on the same 64 outputs per method. Structural alterations count coder changes to the mathematical counterpart, including changes that do not affect the optimum. The repair denominator is the number of incorrect mathematical counterparts.}
\label{tab:coder_audit}
\begin{tabular}{lrr}
\toprule
Audit result & Base LLM & AutoREM \\
\midrule
Incorrect mathematical counterparts & 8/64 (12.5\%) & 1/64 (1.6\%) \\
Objective failures from reformulation errors & 5/64 (7.8\%) & 1/64 (1.6\%) \\
Coder structural alterations & 1/64 (1.6\%) & 2/64 (3.1\%) \\
Coder-only objective failures & 1/64 (1.6\%) & 1/64 (1.6\%) \\
Verified coder repairs & 0/8 & 0/1 \\
Objective accuracy after correcting coder-only errors & 59/64 (92.2\%) & 63/64 (98.4\%) \\
\bottomrule
\end{tabular}
\end{table}


\section{Baseline implementation and offline adaptation protocols}
\label{appx:baseline_and_coverage}


AutoREM, ACE, and ReasoningBank each receive $10\times8=80$ training-sample exposures from the same RO generator and distribution. They use DeepSeek-V4-Flash for reformulation and memory agents, the same fixed DeepSeek-V4-Pro coder, and the same Gurobi solver. Memory construction occurs offline, and all memories are frozen during testing. This matches training exposure and the base models.

\paragraph{ACE.}
The playbook starts with seven empty section headings and no RO rules. Each sample passes through its Generator and the common coder/solver. The Generator--Reflector--Curator mechanism is retained, using three Reflector calls per sample, addition-only curation, and semantic deduplication. At test time, the frozen playbook is supplied to the Generator, followed by the common coder and evaluator. No test-time playbook updates are performed.

\paragraph{ReasoningBank.}
The bank starts empty. When entries are available, each training trajectory retrieves the top-1 entry by cosine similarity, runs the common reformulator--coder--solver pipeline, and uses the original successful- or failed-trajectory induction prompt to append memory. Its accumulation is moved from an online test stream to the offline training split. At test time, retrieval uses the frozen bank and performs no further memory induction.

\paragraph{Formulation-method baselines.}
Base LLM, Max Thinking, Expert Prompt, ReasoningBank, and ACE directly address pure reformulation. AlphaOPT, LEAN-LLM-OPT, and OptiTree primarily address the end-to-end Application task. In their additional pure-reformulation evaluations, the natural-language-to-nominal-formulation step is bypassed: each receives the same \LaTeX\ robust model and is instructed to produce a deterministic robust counterpart. Each retains its original pipeline, and final objectives are evaluated with the same $10^{-4}$ matching tolerance. These comparisons test transfer of formulation-oriented pipelines to algebraic reformulation.

\section{Offline adaptation algorithm}
\label{appx:offline_alg}

Algorithm~\ref{alg:offline} provides the complete pseudocode for the offline adaptation procedure introduced in Section~\ref{sec:offline}.
The procedure iterates over $E$ epochs, each consisting of $S$ training steps.
At each step, a training instance $q_s$ is sampled on the fly and evaluated against the current memory $\mathcal{M}$; if already solved correctly, the step is skipped.
Otherwise, an inner reflection loop of up to $K$ iterations is entered: the \textbf{SMO}-based Reflector proposes structured memory edits (line~11), the \textbf{ULE}-based updater applies them at the granularity of individual uncertain constraint rows (line~12), and the updated memory is re-evaluated on $q_s$.
Once the training instance is solved correctly, the \textbf{Dual-Check Commit (DCC)} mechanism stress-tests the proposed memory against a locked batch $\mathcal{V}_s$ drawn from the validation set (line~16): only if all samples in $\mathcal{V}_s$ pass is the update committed; otherwise the failure feedback is fed back into the next reflection iteration.
If no commit is achieved within $K$ iterations, the candidate memory is discarded and $\mathcal{M}$ remains unchanged.
At the end of each epoch, the \textbf{Validation-based Acceptance (VBA)} gate evaluates the full validation set $\mathcal{V}$ (line~24): an improvement or tie is accepted; if the epoch-level accuracy falls below the best recorded $\mathrm{acc}^*$, the memory is rolled back to $\mathcal{M}^*$, preventing harmful accumulated changes from persisting.
The best memory $\mathcal{M}^*$ is returned after all epochs.

\begin{algorithm}[t]
\caption{Offline adaptation of reformulation experience memory}
\label{alg:offline}
\begin{algorithmic}[1]
\Require Epochs $E$; steps per epoch $S$; DC batch size $B$; max inner iters $K$;
         validation set $\mathcal{V}$; initial memory $\mathcal{M}$
\Ensure Adapted memory $\mathcal{M}^*$
\State $\mathcal{M}^* \leftarrow \mathcal{M}$;\quad $\mathrm{acc}^* \leftarrow \textsc{EvalVal}(\mathcal{M}, \mathcal{V})$
\For{$e = 1, \ldots, E$}
  \For{$s = 1, \ldots, S$}
    \State Sample training instance $q_s$ on the fly
    \State $(e_m, \text{fb}, y) \leftarrow \textsc{Eval}(q_s, \mathcal{M})$ \Comment{Reformulator + Coder + Solver}
    \If{$e_m = 1$} \textbf{continue} \EndIf \Comment{Already correct — skip}
    \State $\hat{\mathcal{M}} \leftarrow \mathcal{M}$;\; $\text{prev\_trace} \leftarrow \varnothing$;\; $\text{dc\_fail} \leftarrow \varnothing$;\; $\mathrm{committed} \leftarrow \text{False}$
    \State \textbf{Lock} DC sample set $\mathcal{V}_s \subseteq \mathcal{V}$, $|\mathcal{V}_s| \leq B$ \Comment{linked-first priority; locked for this step}
    \For{$k = 1, \ldots, K$} \Comment{Inner reflection loop}
      \State \textbf{[\,SMO\,]}\; $\mathrm{ops} \leftarrow \textsc{Reflect}(q_s,\, \hat{\mathcal{M}},\, y,\, \text{fb},\, \text{prev\_trace},\, \text{dc\_fail})$
      \State \textbf{[\,ULE\,]}\; Apply $\mathrm{ops} \in \{\textsc{add}, \textsc{update}, \textsc{delete}\}$ to $\hat{\mathcal{M}}$
      \State $(e_m', \text{fb}', y') \leftarrow \textsc{Eval}(q_s,\, \hat{\mathcal{M}})$;\quad $\text{prev\_trace} \leftarrow (k, \mathrm{ops}, e_m', \text{fb}')$;\quad $\text{dc\_fail} \leftarrow \varnothing$
      \If{$e_m' = 0$}\; $y \leftarrow y'$;\; $\text{fb} \leftarrow \text{fb}'$;\; \textbf{continue} \EndIf
      \State \textbf{[\,Dual-Check Commit\,]}\; $(\mathrm{pass},\, \text{dc\_fail}) \leftarrow \textsc{DualCheck}(\hat{\mathcal{M}},\, \mathcal{V}_s)$
      \If{$\mathrm{pass}$}\; $\mathcal{M} \leftarrow \hat{\mathcal{M}}$;\; $\mathrm{committed} \leftarrow \text{True}$;\; \textbf{break}
      \Else\; $y \leftarrow y'$;\; $\text{fb} \leftarrow \text{fb}'$ \Comment{DC failed — Reflector gets dc\_fail next iter}
      \EndIf
    \EndFor
    \If{$\neg\,\mathrm{committed}$}\; Discard $\hat{\mathcal{M}}$ \EndIf
  \EndFor
  \State \textbf{[\,Validation-based Acceptance\,]}\; $\mathrm{acc} \leftarrow \textsc{EvalVal}(\mathcal{M},\, \mathcal{V})$
  \If{$\mathrm{acc} \geq \mathrm{acc}^*$}\; $\mathcal{M}^* \leftarrow \mathcal{M}$;\; $\mathrm{acc}^* \leftarrow \mathrm{acc}$
  \Else\; $\mathcal{M} \leftarrow \mathcal{M}^*$ \Comment{Restore best memory}
  \EndIf
\EndFor
\State \Return $\mathcal{M}^*$
\end{algorithmic}
\end{algorithm}

\section{Offline adaptation dynamics}
\label{appx:offline_curve}

\begin{figure}[t]
    \centering
    \includegraphics[width=0.5\linewidth]{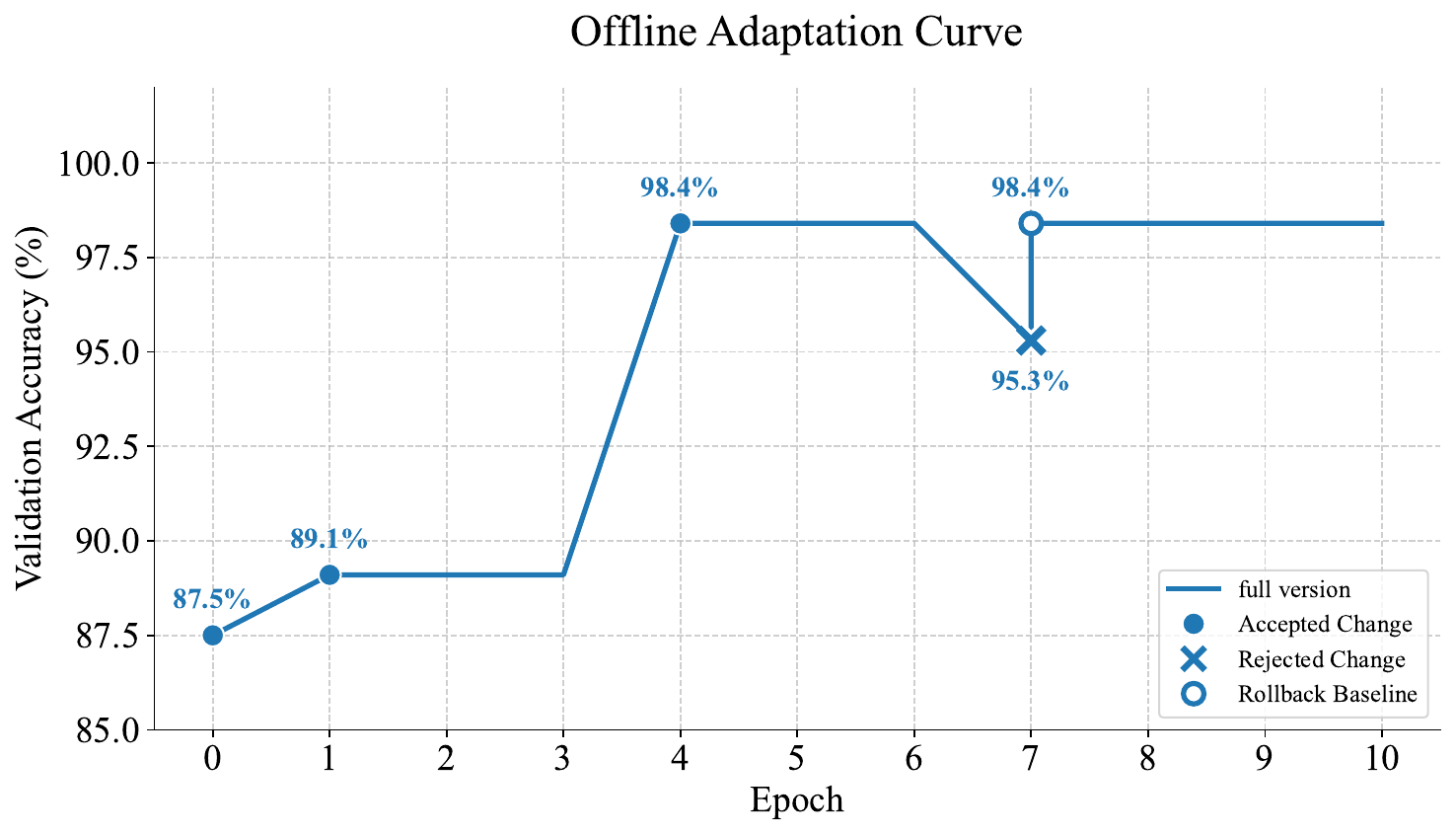}
    \caption{Validation accuracy curve of offline adaptation.}
    \label{fig:offline_curve}
\end{figure}

Figure~\ref{fig:offline_curve} visualizes the evolution of validation accuracy throughout the offline adaptation process. Starting from an initial accuracy of 87.5\%, the model improves rapidly, reaching 89.1\% by epoch 1 and making a decisive leap to 98.4\% by epoch 4. After this point, the accuracy stabilizes at a high plateau. The figure also illustrates the protective role of VBA: at epoch 7, a detrimental memory edit briefly causes validation accuracy to drop to 95.3\%; this update is correctly identified as a rejected change and the rollback baseline is triggered, swiftly restoring accuracy to 98.4\%. This concretely demonstrates that the VBA gate is not merely a passive quality filter but an active safeguard that prevents harmful updates from permanently degrading hard-won improvements. These dynamic trajectories corroborate the static ablation results in Figure~\ref{fig:abla}: both perspectives jointly confirm that all four components of AutoREM make distinct and significant contributions to its performance.

\section{Ablation study details}
\label{appx:ablation}

Each ablation variant removes exactly one component of AutoREM while keeping the rest intact.
\textit{w/o ULE} is implemented by removing unit-level decomposition from the prompts, so the model must process the full constraint matrix as a monolithic block rather than attributing experiences to individual uncertain rows.
\textit{w/o SMO} is implemented by modifying both the prompts and the internal controller to allow only a single addition operator (following \citet{zhang2026agentic}), and replacing operator-aware deduplication with cosine-similarity-based deduplication. This prevents the reflector from updating or deleting existing unit-level entries.
\textit{w/o DCC} skips the dual-check step entirely while retaining epoch-level VBA on the fixed validation set.
\textit{w/o VBA} removes epoch-level validation and rollback. DCC remains active using a local replay buffer, so memory updates are still subject to a step-level check.

As shown in Figure~\ref{fig:abla}, each removal causes a distinct accuracy drop: ULE (97.4\%$\to$90.6\%), SMO (92.2\%), DCC (93.8\%), and VBA (93.8\%).
The largest drop from ULE confirms that unit-level attribution is the backbone of effective memory reuse.
DCC and VBA contribute equally in magnitude but at complementary granularities: DCC catches step-level regressions before they are committed, while VBA prevents epoch-level accumulation of locally beneficial but globally harmful updates.

\section{Hyperparameter sensitivity analysis}
\label{appx:hyperparam}

We study the sensitivity of AutoREM to three key hyperparameters: the dual-check size $B$, the validation set size $|\mathcal{V}|$, and the inner reflection loop cap $K$.

\subsection{Dual-check size \texorpdfstring{$B$}{B}}

Figure~\ref{fig:dc_size} shows the effect of varying $B \in \{1, 2, 4, 8\}$.
Accuracy plateaus at 95.3\% for $B \leq 2$, peaks at \textbf{97.4\%} at $B=4$, then drops to 93.8\% at $B=8$.
The non-monotonic behavior reflects two opposing forces: a larger $B$ improves regression coverage across diverse constraint types, but it also lengthens the failure feedback fed into the Reflector, making it harder to produce a repair that satisfies all sampled instances simultaneously.
$B=4$ balances these two effects.

\subsection{Validation set size \texorpdfstring{$|\mathcal{V}|$}{|V|}}

Figure~\ref{fig:val_size} shows that accuracy rises monotonically from 95.3\% to \textbf{97.4\%} as $|\mathcal{V}|$ grows from 8 to 64, while the generalization gap shrinks from 4.7\% to 1.0\%.
Smaller validation sets introduce high variance into the VBA signal; larger sets improve reliability but increase adaptation cost.
We adopt $|\mathcal{V}|=64$ as the default, where the generalization gap has already stabilized near its minimum.

\subsection{Inner reflection loop cap \texorpdfstring{$K$}{K}}
\label{appx:inner_loop}

We report the empirical usage of the inner reflection loop (Algorithm~\ref{alg:offline}, parameter $K$) on the offline adaptation training run, and use these statistics to assess whether the chosen value $K=3$ is well-calibrated.

\paragraph{Invocation statistics.}
Across the full training run ($E=10$ epochs $\times$ $S=8$ steps $=80$ step attempts), the inner reflection loop was invoked exactly $3$ times, once per training step that was initially incorrect under the current memory. The remaining $77$ steps were already correct on entry and skipped reflection (Algorithm~\ref{alg:offline}, line~6). Table~\ref{tab:inner_loop} lists each invocation with the iterations it consumed and its per-iteration outcomes. The distribution of iterations per invocation is exactly one invocation each at $1$, $2$, and $3$ iterations, with mean $2.0$. Only the first invocation reached the cap $K$, and it terminated by a successful commit at iteration~$3$ rather than by exhausting the budget without a fix.

\begin{table}[t]
\centering
\caption{Per-invocation usage of the inner reflection loop during offline adaptation. Iter outcomes are: \emph{training fail} = the training sample is still incorrect under the proposed memory (no DC performed); \emph{DC fail} = training sample correct but the dual-check batch regressed; \emph{DC pass} = both training sample and DC batch correct (commit).}
\label{tab:inner_loop}
\small
\begin{tabular}{ccl}
\toprule
\textbf{(Epoch, Step)} & \textbf{Iterations Used} & \textbf{Per-iteration outcome} \\
\midrule
$(1, 1)$ & $3$ ($= K$) & Iter~1: training fail; Iter~2: DC fail; Iter~3: DC pass (committed) \\
$(4, 2)$ & $2$         & Iter~1: DC fail; Iter~2: DC pass (committed) \\
$(7, 2)$ & $1$         & Iter~1: DC pass (committed; later rolled back by VBA) \\
\bottomrule
\end{tabular}
\end{table}

\paragraph{Is $K=3$ a reasonable setting?}
The data supports $K=3$ as a tightly calibrated choice for three reasons:
\begin{itemize}
  \item \emph{Large enough to enable the hardest reflection.} The first invocation, which bootstrapped the memory from the empty state, required all $3$ iterations: the very first iteration could not even fix the training sample, the second produced a fix that regressed on the dual-check batch, and only the third yielded a generalizable update that passed both checks. With $K \leq 2$, this foundational update would have been discarded.
  \item \emph{Small enough to control cost.} Only $1$ of the $3$ invocations ($33\%$) reached the cap, and that single cap-hit invocation succeeded at the cap rather than exhausting it. Increasing $K$ to $4$ or beyond would not have changed any outcome or call count on this run, because all invocations already terminate within three iterations; it only increases the permitted worst-case budget on other trajectories.
  \item \emph{Reflection difficulty decays over training.} The three invocations consumed $3$, $2$, and $1$ iterations respectively, indicating that as the memory accumulates correct templates, later reflections converge faster. Beyond epoch~1 the cap is therefore not the binding constraint: it is needed only for the cold-start phase.
\end{itemize}

In summary, $K=3$ provides exactly the headroom required for the hardest observed reflection (cold-start) while remaining tight enough to keep Reflector cost low once the memory has matured.

\section{Qualitative comparison of learned memories}
\label{appx:memory_comparison}

This section provides a qualitative illustration of the knowledge representations learned by AutoREM, ACE, and ReasoningBank, alongside the Expert Prompt used as a baseline.
These examples offer a concrete basis for interpreting the accuracy differences. As shown in Table \ref{tab:mem_size}, AutoREM's memory is \textbf{6$\times$ smaller than ACE} and \textbf{10$\times$ smaller than ReasoningBank}, while achieving strictly higher accuracy. The following subsections illustrate the qualitative contrast between these representations.

\begin{table}[t]
\centering
\caption{Memory size overview.}
\label{tab:mem_size}
\begin{tabular}{lcc}
\toprule
Method & \# Memory Entries & Entry Style \\
\midrule
Expert Prompt & --- & Manually written reformulation rules \\
ReasoningBank & 80 & One item per training step; multi-paragraph narrative \\
ACE           & 51 & 29 strategies + 18 formula entries + 4 code snippets \\
AutoREM (Ours) & \textbf{8} & Compact; one-liner reformulation templates \\
\bottomrule
\end{tabular}
\end{table}

\subsection{AutoREM's learned memory}

AutoREM produces 8 compact entries for observed failure modes, rather than exhaustively enumerating all $3\times2\times2=12$ combinations of uncertainty set type, direction (min/max), and objective/constraint. Memory is learned on demand: reflection is triggered only when the reformulator fails on a training sample. The reformulator prompt requires deriving unmatched units from scratch, so the learned entries supplement observed capability gaps rather than define the full set of supported cases.
Each entry is a self-contained, directly applicable template for one specific combination of uncertainty set type and problem component. There is no redundancy, no contradictory guidance, and no implementation-specific noise. This tight, minimal structure is in stark contrast with the memories learned by ACE and ReasoningBank, as shown below.

A deterministic symbolic reformulator is preferable when the grammar, uncertainty families, and complete correct rules are specified and encoded. AutoREM does not aim to outperform this closed-world compiler. It studies automatic acquisition, revision, deletion, and validation of knowledge from failures and solver feedback, starting from empty memory. Extending a compiler requires experts to derive, implement, and verify new rules. AutoREM targets incomplete or evolving coverage. Expert Prompt instead supplies fixed expert knowledge to an LLM rather than executing symbolic rules.


\lstset{style=promptbox}
\begin{lstlisting}[caption={AutoREM: all 8 learned memory entries.}]
[Entry 1]
For an uncertain objective with box uncertainty:
minimize max_{xi in Box} (c_bar + diag(c_hat) xi)^T x,
where Box = {xi : ||xi||_inf <= 1}.
Reformulate as:
  minimize c_bar^T x + sum_j u_j,
  with u_j >= c_hat_j x_j, u_j >= -c_hat_j x_j
  for all j where c_hat_j > 0.
(u_j are free variables, automatically nonneg by the constraints.)

[Entry 2]
For a constraint with budgeted uncertainty:
min_{xi in Budgeted} (a_bar + diag(a_hat) xi)^T x >= b,
where Budgeted = {xi : ||xi||_inf <= 1, ||xi||_1 <= Gamma}.
Reformulate as:
  a_bar^T x - Gamma*p - sum_j q_j >= b,
  with p >= 0, q_j >= 0,
  p + q_j >= a_hat_j x_j, p + q_j >= -a_hat_j x_j, for each j.

[Entry 3]
For a constraint with polyhedral uncertainty:
min_{delta in Poly} (a_bar + delta)^T x >= b,
where Poly = {delta : D delta <= d, E delta = e}.
Reformulate using LP duality:
  a_bar^T x - d^T lambda - e^T mu >= b,
  D^T lambda + E^T mu = -x, lambda >= 0, mu free.

[Entry 4]
For a constraint with box uncertainty:
min_{xi in Box} (a_bar + diag(a_hat) xi)^T x >= b,
where Box = {xi : ||xi||_inf <= 1}.
Reformulate as:
  a_bar^T x - sum_j w_j >= b,
  with w_j >= 0, w_j >= a_hat_j x_j, w_j >= -a_hat_j x_j,
  for each j where a_hat_j > 0.

[Entry 5]
For an objective with polyhedral uncertainty:
maximize min_{delta in Poly} (c_bar + delta)^T x,
where Poly = {delta : D delta <= d, E delta = e}.
Reformulate as:
  maximize t, subject to
  t <= c_bar^T x - d^T lambda - e^T mu,
  D^T lambda + E^T mu = -x, lambda >= 0, mu free.

[Entry 6]
For a constraint with polyhedral uncertainty and max direction:
max_{delta in Poly} (a_bar + delta)^T x <= b,
where Poly = {delta : D delta <= d, E delta = e}.
Reformulate as:
  a_bar^T x + d^T lambda + e^T mu <= b,
  D^T lambda + E^T mu = x, lambda >= 0, mu free.

[Entry 7]
For a constraint with budgeted uncertainty and max direction:
max_{xi in Budgeted} (a_bar + diag(a_hat) xi)^T x <= b,
where Budgeted = {xi : ||xi||_inf <= 1, ||xi||_1 <= Gamma}.
Reformulate as:
  a_bar^T x + Gamma*p + sum_j q_j <= b,
  with p >= 0, q_j >= 0,
  p + q_j >= a_hat_j x_j, p + q_j >= -a_hat_j x_j, for each j.

[Entry 8]
For an objective with polyhedral uncertainty:
minimize max_{delta in Poly} (c_bar + delta)^T x,
where Poly = {delta : M delta <= q, E delta = e}.
Reformulate as:
  minimize c_bar^T x + q^T lambda + e^T mu,
  subject to M^T lambda + E^T mu = x, lambda >= 0, mu free.
\end{lstlisting}

\subsection{ACE's learned memory (representative samples)}

ACE accumulates 51 entries organized into ``strategies'', ``formulas/calculations'', and ``code snippets''. The strategy entries are verbose, covering overlapping reformulation scenarios with varying levels of detail and occasional sign-convention inconsistencies. We show three representative entries below.

In total, ACE contains 51 entries, many of which are redundant or mutually contradictory on sign conventions (e.g., the max-min stationarity condition is stated differently in \texttt{strat-00088}, \texttt{calc-00030}, \texttt{calc-00089}, and \texttt{calc-00106}). The cross-referencing burden and conflicting signals likely explain why the LLM fails to extract reliable guidance on challenging problem types, contributing to the accuracy drop on the \emph{Hard} dataset.

\begin{lstlisting}[caption={ACE: three representative strategy entries (abridged).}]
[strat-00019]  helpful=169  harmful=0
For robust optimization problems with multiple uncertain components
(objectives and constraints), adopt a modular approach:
(1) Identify each uncertain component and its uncertainty set type
    (box, budget, polyhedral).
(2) For box uncertainty, check variable bounds: if all uncertain
    variables are nonnegative, replace |x_j| with x_j; if nonpositive,
    replace with -x_j; if mixed, use general absolute values and add
    w_j with -w_j <= x_j <= w_j; for budget, introduce p_j and z with
    p_j + z >= hat_j w_j; for polyhedral, dualize with lambda.
(3) Use distinct auxiliary variable names per component (e.g., prefix
    `obj_', `con1_') to avoid accidental coupling.
(4) Include all deterministic constraints and variable bounds unchanged.

[strat-00088]  helpful=44  harmful=3
For systematic dualization of any polyhedral robust component (objective
or constraint), follow this step-by-step Lagrangian approach:
(1) Determine the inner optimization direction: for a <= constraint,
    inner is max; for a >= constraint, inner is min; ...
(2) Write the inner problem as an LP in P delta <= q (convert all
    inequalities including box bounds).
(3) Form the Lagrangian by adding lambda^T(P delta - q) (lambda >= 0)
    and mu^T(E delta - e) (mu unrestricted) ...
(4) Rearrange to isolate delta: Lagrangian = delta^T(x + P^T lambda +
    E^T mu) - lambda^T q - mu^T e ...
... [48 more lines of derivation guidance with sign mnemonic tables] ...

Note: For inner max, stationarity is P^T lambda + E^T mu = x and
penalty is added. For inner min, stationarity is P^T lambda + E^T mu
= -x and penalty is subtracted. Always re-derive the Lagrangian when
uncertain. [Contradicted by strat-00105 which revises this rule.]

[calc-00089]  helpful=50  harmful=1
Correct dualization of polyhedral uncertainty set for all robust
component types:
(1) <= constraint (inner max): a_bar^T x + lambda^T q + mu^T e <= b,
    P^T lambda + E^T mu = x, lambda >= 0, mu unrestricted.
(2) >= constraint (inner min): a_bar^T x - lambda^T q - mu^T e >= b,
    P^T lambda + E^T mu = -x, lambda >= 0, mu unrestricted.
(3) Objective min-max: min c_bar^T x + lambda^T q + mu^T e,
    s.t. P^T lambda + E^T mu = x.
(4) Objective max-min: max c_bar^T x - lambda^T q - mu^T e,
    s.t. P^T lambda + E^T mu = -x.
... [Note: conflicts with calc-00030 and is partially corrected by
calc-00106; sign convention for max-min stationarity is listed
inconsistently across entries.] ...
\end{lstlisting}

\subsection{ReasoningBank's learned memory (representative samples)}

ReasoningBank produces one multi-item memory record per training step (80 in total). Each record contains 2--3 narrative-style memory items grounded in the specific training instance, including concrete numerical values from that instance. We show two representative entries below.

While individual ReasoningBank entries are conceptually sound, the 80-entry corpus is heavily redundant: the same LP-duality insight is restated dozens of times with minor phrasing variations tied to specific training instances. With top-1 retrieval, the LLM reads only the retrieved entry. Redundancy may instead limit the diversity or relevance of retrieved guidance.

\begin{lstlisting}[caption={ReasoningBank: two representative memory items.}]
[epoch1_step1 -- Memory Item 1]
Title: Worst-case direction for box uncertainty in minimization problems
Description: Under box uncertainty and nonneg variables, the worst-case
for minimization over uncertainty is achieved at the extreme point where
each parameter takes the opposite sign of its perturbation coefficient.
Content: For (a_i + b_i xi_i) x_i with xi_i in [-1,1] and x_i >= 0,
the minimum over xi_i occurs at xi_i = -sign(b_i). This rule applies
directly to both robust objectives of the form max min_xi and robust
constraints of the form min_xi (.) >= 0, all with nonnegative variables.

[epoch1_step2 -- Memory Item 1]
Title: Correct dual constraint sign for inner minimization over polyhedral set
Description: When dualizing an inner minimization problem over a
polyhedral uncertainty set in a max-min objective, the dual constraint
must match the negative of the coefficient vector.
Content: In the reformulation of max_x min_{delta: D delta <= d}
(c_bar + delta)^T x, the inner minimization min delta^T x has dual
max_{pi >= 0} -d^T pi subject to D^T pi = -x. A common error is to
write D^T pi = x, which corresponds to the dual of a maximization over
delta. Always derive the dual from the primal direction.

[epoch1_step2 -- Memory Item 3]
Title: Include all indices in dual variables for budgeted uncertainty
Description: When constructing the robust counterpart for box-budget
uncertainty, explicitly include dual variables for every index to
avoid subtle omissions.
Content: Even if some deviation coefficients hat_a_i are zero, the
dual constraint r_i + s >= hat_a_i x_i is still needed. For zero
deviation, the constraint simplifies to r_i + s >= 0, which can be
satisfied by r_i = 0 and s >= 0. Omitting such indices can lead to
an incomplete set of dual variables and potential errors.
\end{lstlisting}

\subsection{Expert Prompt}

The Expert Prompt baseline provides a single fixed prompt containing a set of manually designed reformulation rules that cover the five standard uncertainty set types (ellipsoidal, polyhedral, general $p$-norm ball, 1-norm ball, $\infty$-norm ball). For reference, the core rules are as follows.

\begin{lstlisting}[caption={Expert Prompt: core reformulation rules (from \citet{bertsimas2024robust}).}]
REFORMULATION RULES

Assume a robust constraint (a + P z)^T x <= b  for all z in Z.

- If Z is the Euclidean ball with radius r (||z||_2 <= r):
    a^T x + r ||P^T x||_2 <= b.

- If Z is a polyhedral set (Z = {z : D z <= d}):
    a^T x + d^T y <= b,  D^T y = P^T x,  y >= 0.

- If Z is a general p-norm ball with radius r (||z||_p <= r):
    a^T x + r ||P^T x||_q <= b,
    where 1/p + 1/q = 1  (q is the dual norm of p).

- If Z is a 1-norm ball with radius r (||z||_1 <= r):
    a^T x + r ||x||_inf <= b.

- If Z is an inf-norm ball with radius r (||z||_inf <= r):
    a^T x + r ||x||_1 <= b.
\end{lstlisting}



\section{Generalization across uncertainty structures and problem sources}
\label{appx:transfer-tests}

We construct three datasets to test whether the frozen memory generalizes under changes in uncertainty structure and problem source. They cover non-diagonal perturbation mappings in \emph{General $P_i$}, shared primitive uncertainty across rows in \emph{Coupled Row}, and independently sourced textbook and published robust models in \emph{External}. The \emph{General $P_i$}, \emph{Coupled Row}, and \emph{External} datasets each contain 32 instances. Table~\ref{tab:additional-transfer} reports mean $\pm$ sample standard deviation over three test-time runs.

\begin{table}[t]
\centering
\small
\caption{Generalization results on \emph{General $P_i$}, \emph{Coupled Row}, and \emph{External} with the same frozen memory.}
\label{tab:additional-transfer}
\setlength{\tabcolsep}{4pt}
\begin{tabular}{llrr}
\toprule
Dataset & Method & Accuracy (\%) & Output tokens \\
\midrule
\emph{General $P_i$} & Base LLM & $89.6\pm4.8$ & $8574\pm49$ \\
 & AutoREM & $90.6\pm5.4$ & $8379\pm290$ \\
\emph{Coupled Row} & Base LLM & $76.0\pm4.8$ & $8055\pm183$ \\
 & AutoREM & $89.6\pm4.8$ & $6788\pm328$ \\
\emph{External} & Base LLM & $92.7\pm1.8$ & $5195\pm236$ \\
 & AutoREM & $97.9\pm1.8$ & $7347\pm263$ \\
\bottomrule
\end{tabular}
\end{table}

\subsection{Controlled non-diagonal perturbation matrices}
\label{appx:general-p}

Beyond the \emph{Random} data-generation setting, every uncertain row in \emph{General $P_i$} uses a non-diagonal perturbation matrix while preserving nominal sparsity, so nominal zeros outside the selected support remain unperturbed. The modest accuracy gain is consistent with partial transfer. The current memory was learned from the identity/diagonal perturbation setting used in training. It captures family- and direction-level dualization patterns, but does not explicitly encode the new $P_i^\top x$ mapping, which must largely be derived at inference.

\subsection{Shared primitive uncertainty across two rows}
\label{appx:coupled-row}

Beyond the \emph{Random} data-generation setting, two uncertain rows in \emph{Coupled Row} share one primitive uncertainty vector and set. AutoREM improves accuracy from 76.0\% to 89.6\% and reduces output tokens by 15.7\%. This is consistent with the familiar row-wise transformations remaining applicable. The new challenge is recognizing and consistently reusing the shared uncertainty set.

\subsection{A literature-derived 32-instance challenge set}
\label{appx:external-ro32}

The \emph{External} dataset contains 32 numerical robust instances from five textbook and published-benchmark sources, independent of the AutoRO-Bench generator. It spans seven problem families: drug production, single-period portfolio, two-variable robust LP, mixed 0--1 production planning, multiperiod production planning, robust selection, and NETLIB robust LP. Table~\ref{tab:external-sources} summarizes the sources.

\begin{table}[t]
\centering
\small
\caption{Sources and problem families of the \emph{External} dataset.}
\label{tab:external-sources}
\begin{tabularx}{\linewidth}{@{}l r X@{}}
\toprule
Source & Instances & Problem families \\
\midrule
\citet{BenTal2009} & 4 & Drug production and single-period portfolio. \\
\citet{li2011robustcounterpart} & 12 & Two-variable robust LP and mixed 0--1 production planning. \\
\citet{li2012probabilistic} & 4 & Multiperiod production planning. \\
\citet{goerigk2024benchmarking} & 6 & Robust selection. \\
\citet{ben-tal2000robust} & 6 & NETLIB robust LP. \\
\bottomrule
\end{tabularx}
\end{table}

The original sources provide numerical robust instances, model equations, or benchmark files. For each instance, we extract the source-defined objective, constraints, variable domains, uncertainty set, and numerical parameters and convert the original robust model into a self-contained \LaTeX\ input. Thus, \emph{External} is evaluated as an RO reformulation task: the explicit robust model is passed directly to the reformulator and checked by the same fixed coder and solver, without the natural-language formulation stage used for the RO application task. The set contains 20 LPs and 12 MILPs; 14 box, eight budget, seven polyhedral, and three box--polyhedral intersection instances. Decision-variable counts range from 2 to 201, original constraint counts from 1 to 56, and uncertain-row counts from 1 to 14. 16 instances have multiple uncertain rows.

With the existing memory frozen and without adaptation to these instances, AutoREM improves accuracy from 92.7\% to 97.9\%, a gain of 5.2\%, while output tokens increase from 5,195 to 7,347.

\section{Additional experiments: DeepSeek-V3.2 as base LLM}
\label{appx:v32}

\textbf{Experimental setup.}
This section reports results on the \emph{Random} dataset using DeepSeek-V3.2 (Chat Mode) as the base LLM, following the same evaluation protocol as the main experiments.
Both the reformulator and the reflector in AutoREM use this base LLM, and offline adaptation is performed on DeepSeek-V3.2 (Chat Mode) from scratch.
Note that the Max Thinking baseline here refers to DeepSeek-V3.2 with Thinking Mode.
The evaluation is conducted on the \emph{Random} dataset, using 32 instances in a single run.

\textbf{Results and analysis.}
Table~\ref{tab:result-v3.2} reports the results.
The base LLM achieves only 46.9\% accuracy, a drop of over 40 percentage points from the DeepSeek-V4-Flash result of 87.5\% in Table~\ref{tab:main}, reflecting V3.2's considerably weaker capability on this algebraically demanding task.
Baselines improve upon the base LLM only partially. Max Thinking reaches 59.4\%, Expert Prompt 68.8\%, and ACE (the strongest baseline) 71.9\%, all far below their V4-Flash counterparts, indicating that a weaker base LLM propagates degradation uniformly across the compared baselines.

\begin{table}[h]
\centering
\caption{Comparison of methods with DeepSeek-V3.2 as the base LLM.}
\label{tab:result-v3.2}
    \begin{tabular}{lcc}
    \toprule
    \multirow{2}[2]{*}{Methods} & \multicolumn{2}{c}{Random} \\
          & Accuracy/\%↑ & Output Tokens↓ \\
    \midrule
    Base LLM & 46.9  & 4748 \\
    Max Thinking & 59.4  & 13971 \\
    Expert Prompt & 68.8  & 3734 \\
    ReasoningBank & 53.1  & \underline{2510} \\
    ACE   & \underline{71.9} & 3184 \\
    AutoREM & \textbf{96.9} & \textbf{2006} \\
    \bottomrule
    \end{tabular}%
\end{table}

AutoREM achieves 96.9\% accuracy, nearly identical to its V4-Flash result of 97.4\%, and notably surpasses the V4-Flash base LLM (87.5\%) despite operating on a substantially weaker model.
The absolute gain over the V3.2 base LLM is 50.0\%, far exceeding the ${\approx}$10\% improvement observed on V4-Flash.
The margin over the best baseline (ACE, 71.9\%) reaches 25.0\%, compared to only 4.7 percentage points on V4-Flash.
These results show that the improvement brought by AutoREM is far greater when the base LLM is weaker, suggesting that the offline adaptation procedure is especially effective in filling the larger gap left by a less capable model.
AutoREM also achieves the lowest output token count (2006 tokens), far below Max Thinking (13971 tokens), further showing that AutoREM is both more accurate and more efficient than inference-time computation scaling.

\textbf{Note on DeepSeek-V3.2.}
DeepSeek-V3.2 is the previous-generation flagship model from DeepSeek.
During our experiments, DeepSeek released a new model version and deprecated the V3.2 API, making it impossible to continue experiments on V3.2.
We therefore switched to the latest DeepSeek-V4-Flash for our main experiments, which constitutes the primary results reported in the paper.
The V3.2 results above are included for reference, reflecting the initial stage of our experimental process.

\section{Discussion of empirical scope and security}
\label{appx:scope_and_security}

\subsection{Empirical scope}
\label{appx:scope_boundary}

Our empirical evaluation focuses on static robust linear models with box, budget, and polyhedral uncertainty, including application instances with natural-language problem description. It does not establish performance on ellipsoidal uncertainty, conic or semidefinite counterparts, or adjustable robust optimization. Extending AutoREM to these settings would require suitable data, reformulation units, solvers, and verifiers. Frozen memory may offer limited benefit on entirely unseen structures.

\subsection{Security considerations}
\label{appx:security}

The experiments assume trusted generated or expert-curated inputs and do not evaluate adversarial robustness. Prompt injection, memory poisoning, malformed uncertainty sets, and unsafe generated code remain potential threats. Existing parsing and memory-update controls do not establish resistance to these threats or guarantee safe execution. Deployment with untrusted inputs would require additional semantic validation and sandboxed execution, and high-stakes applications would require expert verification of the resulting model.

\section{LLM prompts}
\label{appx:prompts}

This section gives the verbatim prompts used for the three LLM agents in AutoREM.
Dynamic inputs inserted at runtime are marked with \texttt{<<PLACEHOLDER>>} notation.

\subsection{Reformulator prompt}
\label{appx:prompts_reformulator}

\begin{lstlisting}[style=promptbox]
You are a Reformulator.

Your task is to convert the input robust optimization problem into a tractable
robust counterpart -- a fully explicit deterministic formulation in which all
effects of uncertainty are resolved, with no shorthand operators remaining
(e.g., absolute values |*|, max/min, norms, or supremum expressions).

Use the experience memory to assist: match each minimal independent reformulation
unit (e.g., a single uncertain constraint or uncertain objective term) to the
most relevant experience if one exists, and apply that experience's reformulation
approach; derive the counterpart from scratch for unmatched units.

All specific numeric values from the problem instance must be preserved and
appear explicitly in the final answer.

==================================================================
INPUT
==================================================================

=== PROBLEM ===
<<PROBLEM INPUT>>

=== REFORMULATION EXPERIENCE MEMORY ===
<<EXPERIENCE MEMORY>>

==================================================================
OUTPUT FORMAT
==================================================================

Your ENTIRE response MUST be a single valid JSON object -- nothing else.
  * Do NOT include any text, commentary, or markdown outside the JSON.
  * Do NOT wrap the JSON in code fences.
  * The FIRST character of your response MUST be { and the LAST MUST be }.

"reasoning" MUST cover:
  (a) For each reformulation unit: which experience was matched (or none) and why.
  (b) The derivation process that transforms each unit into its tractable
      robust counterpart.

"matched_experience_ids" MUST be a list of integer experience IDs matched to
reformulation units in (a). Use [] if no unit was matched.

"final_answer" MUST be a complete mathematical formulation of the tractable
robust counterpart in LaTeX within a \begin{align}...\end{align} environment,
with all numeric values from the problem instance appearing explicitly.

Return ONLY a RAW JSON object with the following structure:
{
  "reasoning": <matching analysis and derivation process for each reformulation unit>,
  "matched_experience_ids": [<list of matched experience IDs as integers>],
  "final_answer": <the tractable robust counterpart with numeric values in LaTeX>
}
\end{lstlisting}

\subsection{Coder prompt}
\label{appx:prompts_coder}

\begin{lstlisting}[style=promptbox]
You are a Coder.

Your task is to write Python code that uses the `gurobipy' library to solve
the given optimization problem (a tractable robust counterpart).

Import gurobipy and create a Model for the given problem. Solve the model, and
then print the optimal objective value using print(model.ObjVal). If the model
is infeasible or unbounded, print "infeasible" or "unbounded" accordingly.

Note: In Gurobi, the default lower bound for variables is 0. Free variables
(unbounded) MUST be declared explicitly with a negative infinity lower bound
(e.g., lb=float(`-inf') or lb=-GRB.INFINITY).

==================================================================
INPUT
==================================================================

=== PROBLEM (ROBUST COUNTERPART) ===
<<ROBUST COUNTERPART INPUT>>

==================================================================
OUTPUT FORMAT
==================================================================

Your ENTIRE response MUST be a single valid JSON object -- nothing else.
  * Do NOT include any text, commentary, or markdown outside the JSON.
  * Do NOT wrap the JSON in code fences.
  * The FIRST character of your response MUST be { and the LAST MUST be }.

Return ONLY a RAW JSON object with the following structure:
{
  "reasoning": <step-by-step thinking on how to translate the formulation into code>,
  "final_answer": <the complete Python code>
}
\end{lstlisting}

\subsection{Reflector prompt}
\label{appx:prompts_reflector}

\begin{lstlisting}[style=promptbox]
You are a Reflector.

Your task is to reflect on all training samples in the current step (both
correct and incorrect) and update the reformulation experience memory -- a
collection of experiences that each capture how a specific type of robust
optimization structure should be correctly reformulated into its tractable
robust counterpart. Use three operators: `add', `update', and `delete'.
Each experience_id MUST appear at most once across all operators.

The usage scores shown in the memory (success / failure counts) record how
often each experience has led to a correct or incorrect final answer when used.

Note: A problem may match multiple experiences (one per reformulation unit).
A failure does not imply every matched experience is wrong -- the root cause
may lie in only one of the matched experiences, while others are correct.

In your `reasoning', analyze the root cause of each failure by examining the
Reformulator's matched experience IDs, its reasoning, and the environment
feedback for all samples. Use the correct samples as reference to understand
what works, and focus your operator decisions on resolving the failures without
disrupting what is already correct. Consider all samples together and decide on
a unified set of memory operations. Then decide which operators to apply:
  - `add' a new experience if no suitable one exists in memory.
  - `update' an existing experience if its content is incorrect or misleading.
  - `delete' an experience if it is harmful.

==================================================================
INPUT
==================================================================

=== EXISTING REFORMULATION EXPERIENCE MEMORY ===
<<EXPERIENCE MEMORY>>

--- SAMPLE 1 (CORRECT / INCORRECT) ---

=== PROBLEM ===
<<PROBLEM INPUT>>

=== REFORMULATOR'S OUTPUT ===
<<REFORMULATOR OUTPUT>>

=== ENVIRONMENT FEEDBACK ===
<<SOLVER FEEDBACK>>

<<... ADDITIONAL SAMPLES ...>>

==================================================================
OUTPUT FORMAT
==================================================================

Your ENTIRE response MUST be a single valid JSON object -- nothing else.
  * Do NOT include any text, commentary, or markdown outside the JSON.
  * Do NOT wrap the JSON in code fences.
  * The FIRST character of your response MUST be { and the LAST MUST be }.

Fields per operator:
- add:    {"operator": `add', "experience_id": int, "content": "..."}
- update: {"operator": `update', "experience_id": int, "content": "..."}
- delete: {"operator": `delete', "experience_id": int}

Note on `add' experience_id: assign max(existing experience ids) + 1 for the
first `add', max(existing experience ids) + 2 for the second, and so on.
Each `add' MUST use a distinct new ID.

Return ONLY a RAW JSON object with the following structure:
{
  "reasoning": <step-by-step thinking covering: (1) failure diagnosis,
                (2) correct reformulation derivation for the affected unit(s),
                and (3) operator decisions>,
  "final_answer": [
    { ... }
  ]
}
\end{lstlisting}


\clearpage
\section*{NeurIPS Paper Checklist}

\begin{enumerate}

\item {\bf Claims}
    \item[] Question: Do the main claims made in the abstract and introduction accurately reflect the paper's contributions and scope?
    \item[] Answer: \answerYes{}
    \item[] Justification: The abstract and introduction clearly state the four contributions: AutoRO-Bench benchmark, AutoREM framework, the offline adaptation procedure, and the experimental validation. All claims are supported by the experimental results.

\item {\bf Limitations}
    \item[] Question: Does the paper discuss the limitations of the work performed by the authors?
    \item[] Answer: \answerYes{}
    \item[] Justification: The Conclusion section explicitly discusses two limitations: (1) offline adaptation relies on solver-verified objective values, which may not be available for all problem classes; (2) the framework currently handles only LP-based robust counterparts and does not extend to more complex RO problem classes.

\item {\bf Theory assumptions and proofs}
    \item[] Question: For each theoretical result, does the paper provide the full set of assumptions and a complete (and correct) proof?
    \item[] Answer: \answerNA{}
    \item[] Justification: This paper does not include formal theoretical results (theorems or lemmas). The mathematical derivations of robust counterparts are provided in the appendix as standard reference material, not as novel theoretical contributions.

    \item {\bf Experimental result reproducibility}
    \item[] Question: Does the paper fully disclose all the information needed to reproduce the main experimental results of the paper to the extent that it affects the main claims and/or conclusions of the paper (regardless of whether the code and data are provided or not)?
    \item[] Answer: \answerYes{}
    \item[] Justification: The paper reports all hyperparameters ($E$, $S$, $B$, $K$, $|\mathcal{V}|$), base LLM choices, decoding settings, and evaluation protocols. The data generation pipeline is described in details. Code and datasets will be publicly released.

\item {\bf Open access to data and code}
    \item[] Question: Does the paper provide open access to the data and code, with sufficient instructions to faithfully reproduce the main experimental results, as described in supplemental material?
    \item[] Answer: \answerYes{}
    \item[] Justification: The paper states that datasets and code will be publicly released to support reproducibility. The automated data generation pipeline further allows independent regeneration of training and evaluation data.

\item {\bf Experimental setting/details}
    \item[] Question: Does the paper specify all the training and test details (e.g., data splits, hyperparameters, how they were chosen, type of optimizer) necessary to understand the results?
    \item[] Answer: \answerYes{}
    \item[] Justification: Section~5 specifies all experimental settings including dataset descriptions, baseline details, hyperparameter values with justification, evaluation metrics, and the fixed coder LLM. Additional implementation details are provided in the appendix.

\item {\bf Experiment statistical significance}
    \item[] Question: Does the paper report error bars suitably and correctly defined or other appropriate information about the statistical significance of the experiments?
    \item[] Answer: \answerYes{}
    \item[] Justification: All methods are run three independent times and results are reported as mean $\pm$ standard deviation. The source of variability (engineering-level non-determinism in API calls under greedy decoding) is explicitly stated.

\item {\bf Experiments compute resources}
    \item[] Question: For each experiment, does the paper provide sufficient information on the computer resources (type of compute workers, memory, time of execution) needed to reproduce the experiments?
    \item[] Answer: \answerYes{}
    \item[] Justification: All LLMs are accessed via commercial APIs; no local GPU resources are required. The primary compute cost is in API token consumption, which is characterized by the output token counts reported in Section 5.
    
\item {\bf Code of ethics}
    \item[] Question: Does the research conducted in the paper conform, in every respect, with the NeurIPS Code of Ethics \url{https://neurips.cc/public/EthicsGuidelines}?
    \item[] Answer: \answerYes{}
    \item[] Justification: The research involves automated optimization reformulation with no ethical concerns. All datasets are synthetically generated or curated from publicly available benchmarks with proper attribution, and no sensitive personal data is used.

\item {\bf Broader impacts}
    \item[] Question: Does the paper discuss both potential positive societal impacts and negative societal impacts of the work performed?
    \item[] Answer: \answerNA{}
    \item[] Justification: This work addresses automated mathematical reformulation for robust optimization, a foundational research topic with no foreseeable direct negative societal impacts. The positive impact is democratizing access to robust optimization by removing the manual reformulation bottleneck.
    
\item {\bf Safeguards}
    \item[] Question: Does the paper describe safeguards that have been put in place for responsible release of data or models that have a high risk for misuse (e.g., pre-trained language models, image generators, or scraped datasets)?
    \item[] Answer: \answerNA{}
    \item[] Justification: The released assets (benchmark datasets and optimization framework code) pose no risk of misuse; they are domain-specific tools for mathematical reformulation with no dual-use concerns.

\item {\bf Licenses for existing assets}
    \item[] Question: Are the creators or original owners of assets (e.g., code, data, models), used in the paper, properly credited and are the license and terms of use explicitly mentioned and properly respected?
    \item[] Answer: \answerYes{}
    \item[] Justification: All external assets are properly cited: RSOME, IndustryOR, OptiBench, and all baseline methods. All LLMs are accessed through official commercial APIs under their respective terms of service.

\item {\bf New assets}
    \item[] Question: Are new assets introduced in the paper well documented and is the documentation provided alongside the assets?
    \item[] Answer: \answerYes{}
    \item[] Justification: AutoRO-Bench is documented in Section~4 with full generation pipeline details. The AutoREM framework is described in Section~4 and the appendix. Both the dataset and code will be released publicly with documentation to support reproducibility.

\item {\bf Crowdsourcing and research with human subjects}
    \item[] Question: For crowdsourcing experiments and research with human subjects, does the paper include the full text of instructions given to participants and screenshots, if applicable, as well as details about compensation (if any)?
    \item[] Answer: \answerNA{}
    \item[] Justification: The paper does not involve crowdsourcing. The expert verification of the RO application dataset was conducted by three domain experts as a quality check, not as a formal human subjects study involving data collection from participants.

\item {\bf Institutional review board (IRB) approvals or equivalent for research with human subjects}
    \item[] Question: Does the paper describe potential risks incurred by study participants, whether such risks were disclosed to the subjects, and whether Institutional Review Board (IRB) approvals (or an equivalent approval/review based on the requirements of your country or institution) were obtained?
    \item[] Answer: \answerNA{}
    \item[] Justification: The paper does not involve research with human subjects; IRB approval is not required.

\item {\bf Declaration of LLM usage}
    \item[] Question: Does the paper describe the usage of LLMs if it is an important, original, or non-standard component of the core methods in this research? Note that if the LLM is used only for writing, editing, or formatting purposes and does \emph{not} impact the core methodology, scientific rigor, or originality of the research, declaration is not required.
    \item[] Answer: \answerYes{}
    \item[] Justification: LLMs are a core component of the proposed method. AutoREM uses LLMs in two roles: as the reformulator that generates robust counterparts and as the reflector that proposes memory edits during offline adaptation. All LLM choices and configurations are described in Section~5.

\end{enumerate}
\end{document}